\PassOptionsToPackage{nomath}{stix2}
\documentclass[ASNA,twocolumn]{USG}
\usepackage{anyfontsize}
\usepackage{colortbl}
\usepackage{xcolor}
\usepackage{steinmetz}
\usepackage{amsmath}
\usepackage{booktabs}
\usepackage{graphicx}
\usepackage{adjustbox}
\usepackage{array}
\usepackage{multirow}
\usepackage{rotating}
\usepackage{afterpage}

\usepackage{refcount}
\AddToHook{shipout/background}{%
  \ifnum\value{page}=\getpagerefnumber{tab:faults}\relax
  \fi
}

\usepackage{textcomp}
\usepackage{etoolbox}
\usepackage{unicode-math}
\makeatletter
\patchcmd{\@maketitle}{\includegraphics[width=56mm]{allergy-eps-converted-to.pdf}}{}{}{}
\patchcmd{\@maketitle}{\includegraphics[width=26.5mm]{Wiley_logo-eps-converted-to.pdf}}{}{}{%
  \PackageError{manuscript}{Wiley logo patch failed}{Check the title definition in USG.cls.}}
\patchcmd{\oddfoot@titlepage@info}
  {{\it \@journal}, \@copyyear;v\@volume:\@FirstPg--\@LastPg}
  {{\it \@journal}}
  {}{\PackageError{manuscript}{Preprint footer patch failed}{Check the footer definition in USG.cls.}}
\patchcmd{\oddfoot@titlepage@info}
  {\href{https://doi.org/\thearticledoi}{https://doi.org/\thearticledoi}}
  {\mbox{}}{}{\PackageError{manuscript}{DOI footer patch failed}{Check the footer definition in USG.cls.}}
\patchcmd{\@maketitle}{\vskip24.36mm}{\vskip18mm}{}{}

\patchcmd{\@@articletype}{\raisebox{-2\@p@t}{\includegraphics[width=0.70in]{images/openaccess-eps-converted-to.pdf}}}{}{}{\PackageError{manuscript}{Open Access badge patch failed}{Check the article-type definition in USG.cls.}}
\makeatother

\DeclareGraphicsExtensions{.pdf,.png,.jpg,.jpeg,.eps}
\graphicspath{{./images/}{./images/livo_riskmap/}}

\specialissue{}
\articletype{ORIGINAL ARTICLE}
\subarticletype{}

\journal{Preprint}
\volume{0}
\copyyear{2026}
\startpage{1}
\articledoi{}

\begin{document}

\title{QVS-LIVO RiskMap: Health- and Information-Aware View Selection for Multi-Camera LIVO and Off-Road Risk Mapping}
\transtitle{QVS-LIVO RiskMap: Health- and Information-Aware View Selection for Multi-Camera LIVO and Off-Road Risk Mapping}
\subtranstitle{}

\author[1]{Zihan Pan}[https://orcid.org/0009-0003-0277-6523]
\author[1]{Jinghan Wang}[https://orcid.org/0009-0004-9977-1848]
\author[1]{Hang Li}
\author[1]{Feng Cheng}[https://orcid.org/0000-0003-3726-1821]
\author[1]{Tianchen Liu}[https://orcid.org/0000-0002-9070-978X]
\author[1]{Gaoliang Peng}[https://orcid.org/0000-0002-8543-9455]
\address[1]{State Key Laboratory of Robotic Technology and System, Harbin Institute of Technology, Harbin 150001, China}

\corres{Tianchen Liu (\email{tianchenliu@hit.edu.cn})}

\newcommand{\authoremailinfo}{%
  \textbf{Author emails:}\par
  Zihan Pan: 23b908013@stu.hit.edu.cn\par
  Jinghan Wang: 23b908011@stu.hit.edu.cn\par
  Hang Li: lihang@hit.edu.cn\par
  Feng Cheng: 20250313@hit.edu.cn\par
  Tianchen Liu: tianchenliu@hit.edu.cn\par
  Gaoliang Peng: pgl7782@hit.edu.cn\par
}
\makeatletter
\patchcmd{\printabstractpart}{\ifx\@corresinfo\empty}{%
  {\reset@font\FIfont\authoremailinfo\par}\vskip4.25\@p@t
  \ifx\@corresinfo\empty}{}{\PackageError{manuscript}{Author email fields patch failed}{Check printabstractpart in USG.cls.}}
\makeatother

\historydates{}

\fundingInfo{National Natural Science Foundation of China under Grant No. 52275099, Self-Planned Task of State Key Laboratory of Robotics and Systems (HIT) No.SKLRS202605C}

\titlemark{QVS-LIVO RiskMap}

\keywords{off-road robotics | multi-camera LIVO | semantic mapping | traversability | dual-scale BEV risk map}

\abstract[ABSTRACT]{
Multi-camera observations in off-road environments vary significantly in reliability, posing challenges when complementing degraded geometric constraints. Motivated by this, we propose Quality-aware View Selection Lidar-Vision Odometry Risk Mapping (QVS-LIVO RiskMap), which decouples visual processing for state estimation from that for semantic mapping. The front end evaluates camera observations through read-only pretracking on a shared lidar–inertial posterior, combining observation health with geometric complementarity. Channel health memory and switching hysteresis govern selection of at most one eligible view for direct photometric correction. The posterior is retained when no view is eligible or the correction fails acceptance. The mapping back end integrates semantic observations from all available views in a shared world grid. Vehicle-scale geometry constrains semantic risk corrections, while spatial and temporal evidence supports risk confirmation. The system also produces a local bird's-eye-view (BEV) risk map and a global 2.5D semantic-terrain risk map with obstacle instance voxels. On two field sequences, QVS-LIVO achieved horizontal root mean square errors of 0.910 m and 1.664 m. It reduced error by 43.3\% relative to FAST-LIO2 on the degraded long-route sequence. Across 12 controlled camera-fault configurations involving dropout, black frames and 60 ms timestamp shifts, it maintained complete trajectories with errors of 1.54–2.23 m. When deployed onboard an NVIDIA Jetson AGX Orin, the system generates local and global maps at 9.94 Hz and 2.00 Hz, respectively, with a median local mapping latency of 32.4 ms. These runtime results demonstrate the real-time feasibility of selective visual state updates alongside multi-view semantic mapping under challenging off-road conditions. The code and related resources are available at:
\url{https://github.com/zihanPan-HIT/QVS-LIVO}.
}

\abbr{ATE, absolute trajectory error, BEV, bird's-eye view, GNSS, global navigation satellite system, LIO, lidar--inertial odometry, LIVO, lidar--inertial--visual odometry, QVS-LIVO, quality-aware view-selective lidar--inertial--visual odometry, RMSE, root mean square error, RPE, relative pose error, RTK, real-time kinematic.}

\maketitle

\section{Introduction}

\subsection{Motivation}

Off-road mobile robots performing autonomous inspection, search and rescue, and environmental reconstruction in forests, grasslands and unstructured terrains must continuously estimate their poses and motion with limited prior map information while assessing, in real time, the traversability risks posed by the surrounding terrain. Unlike structured roads, off-road environments lack clear lane markings and flat roadbeds. Pronounced terrain variations, intermittent canopy occlusion and abrupt illumination changes can degrade geometric or visual observations during motion. These conditions disrupt the continuity of features used for motion estimation and increase ambiguity in surface-material recognition and traversability-risk assessment. These challenges motivate robust state estimation under sensor degradation. They also motivate traversability representations that integrate terrain geometry and appearance while accounting for vehicle traversal capabilities \citep{Beycimen2026TerrainTraversability,Frey2024RoadRunner}.

Heterogeneous sensor fusion provides a physical foundation to address these challenges: lidar supplies accurate metric geometry and structural scale, cameras capture rich surface textures and semantic appearance, and the inertial measurement unit (IMU) provides high-frequency motion priors and inertial constraints. However, combining these modalities does not ensure sufficient measurement constraints throughout operation. Insufficient geometric structure can weaken lidar pose constraints along specific degrees of freedom \citep{Liao2025Degeneracy}. Motion blur, occlusions, and lighting changes can make visual tracking unreliable \citep{Cao2026OmniLIVO}. Surround-camera configurations substantially expand the field of view, but image quality and contributions to pose constraints vary considerably across viewing directions, compounded by asynchronous multi-camera arrivals. On resource-constrained edge platforms, identifying and selecting visual observations that provide reliable and complementary constraints along weakly constrained lidar pose directions from asynchronous multi-view streams is a key challenge in containing error propagation and leveraging camera redundancy.

State estimation and semantic mapping also impose different requirements on multi-camera observations. For state estimation, incorporating degraded views or observations with incorrect temporal associations increases computational cost and risks corrupting the pose estimate. For environmental mapping, an image excluded from the current pose update may still capture surfaces or obstacles outside the primary field of view, conversely, a single view that supports stable visual tracking may not cover all terrain requiring assessment. This distinction motivates two complementary design objectives. The front end should evaluate observation health and geometric complementarity on a shared lidar--inertial posterior to govern visual updates. It should retain that posterior when no view is eligible or an update is rejected. Independently, the mapping pipeline should integrate semantic observations from all available views while enforcing vehicle-scale geometric constraints and spatiotemporal risk confirmation.

\subsection{Literature Review}

Prior research relevant to these challenges spans three areas: multimodal localization under geometric degeneracy, multi-camera temporal coordination, and geometric--semantic terrain mapping.

\textbf{(1) Multimodal localization and constraints under degeneracy.} Tightly coupled LIVO combines geometric, photometric and inertial constraints for state estimation. LVI-SAM couples visual--inertial and lidar--inertial subsystems through factor-graph optimization \citep{Shan2021LVISAM}. R3LIVE and FAST-LIVO combine lidar--inertial estimation with direct photometric alignment \citep{Lin2022R3LIVE,Zheng2022FASTLIVO}. FAST-LIVO2 sequentially applies lidar and visual updates within an error-state iterated Kalman filter (ESIKF), using point-to-plane registration and image-patch alignment \citep{Zheng2025FastLIVO2}. Omni-LIVO extends this approach to synchronized surround cameras through cross-view photometric transfer, joint multi-view updates and adaptive covariance inflation \citep{Cao2026OmniLIVO}. Asynchronous camera arrivals and limited computing resources motivate selective visual updates. Candidate selection and update acceptance must account for observation reliability before modifying the current lidar--inertial posterior.

\textbf{(2) Multi-camera observation utilization and asynchronous coordination.} Multi-camera systems provide redundancy against visual degradation confined to individual viewing directions. MIMC-VINS fuses asynchronous camera and IMU measurements through on-manifold state interpolation and maintains estimation under sensor failures \citep{Eckenhoff2021MIMCVINS}. MAVIS extends visual tracking and visual--inertial optimization to multiple cameras \citep{Wang2024MAVIS}. VINS-Multi uses dynamic feature quotas and frame-priority coordination for asynchronous camera streams \citep{Wang2024VINSMulti}, while ARTI-VIO combines feature recall with time-interpolated optimization \citep{Feng2026ARTIVIO}.

Information-aware visual cue selection and active feature matching balance estimation accuracy and computational cost \citep{Carlone2019Attention,Zhao2020GoodFeature}. For multi-camera VIO, Zhang et al. maintain a fixed total feature budget across cameras while jointly estimating motion \citep{Zhang2022BalancingBudget}.

Observation health and geometric complementarity describe distinct properties of a visual observation. Eigenstructure-based degeneracy analysis identifies uneven constraints across pose directions \citep{Zhang2016Degeneracy,Liao2025Degeneracy}. Photometric residuals, tracked point counts and image freshness characterize observation health, but do not directly measure visual support along weakly constrained lidar directions. A healthy view may reinforce directions already well constrained by lidar, while a complementary view may be unreliable. We therefore evaluate both properties on a shared LIO posterior and retain that posterior when no view is eligible or a correction is rejected.

\textbf{(3) Off-road risk mapping through semantic and geometric fusion.} Terrain geometry provides measures of slope, step height, roughness and obstacle structure for traversability assessment. Probabilistic elevation mapping accounts for localization and range-measurement uncertainty \citep{Fankhauser2018ProbabilisticTerrainMapping}, while GPU-based elevation mapping supports efficient terrain processing \citep{Miki2022ElevationGPU}. Forkel et al. estimate terrain surfaces separately from obstacles \citep{Forkel2021TerrainEstimation}. Xue et al. combine multiple lidar frames with spatiotemporal inference for terrain modeling and traversability analysis \citep{Xue2023TerrainModeling}.

Visual terrain-property prediction provides complementary appearance information \citep{Wellhausen2019WhereShouldIWalk}. Multilayer Bayesian mapping jointly represents semantic occupancy and traversability \citep{Gan2023MultilayerMapping}. Beycimen et al. fuse appearance classifications and geometric statistics in a 2.5D traversability grid \citep{Beycimen2026TerrainTraversability}. RoadRunner learns elevation and traversability from camera and lidar features in a unified BEV representation \citep{Frey2024RoadRunner}.

Combining geometric and semantic observations in unstructured field navigation introduces two practical challenges. The first is asymmetry in decision authority. Appearance semantics is susceptible to misclassification caused by domain shift, shadows, or debris (e.g., misclassifying a low obstacle as vegetation associated with a low appearance-risk prior). %In our formulation, appearance-based corrections remain bounded, and cannot reduce the risk assigned to cells classified as direct hard obstacles. 
The second challenge is spatiotemporal risk confirmation. Transient classification flicker and point-cloud edge outliers can produce false alarms. Risk confirmation should distinguish direct obstacle evidence from slope- or step-induced candidates that require additional spatial and temporal support. Furthermore, restricting semantic mapping to the cameras selected for pose estimation would limit environmental coverage; the mapping pipeline should therefore ingest semantic streams across all available views independently.

These considerations motivate coordinating two distinct uses of multi-camera observations in our framework. Through multi-view pretracking, the localization front end evaluates observation health and directional geometric complementarity to select at most one eligible view for visual correction, falling back to the current LIO posterior when updates are unavailable or rejected. Concurrently, the mapping back end aggregates semantic streams across all available views under explicit geometric constraints and spatiotemporal confirmation. Anchored to a shared state reference, this separation limits the visual correction workload while retaining semantic coverage from all available views.

\subsection{Main Contributions}
\label{sec:contributions}

To address the robust perception and mapping for off-road environments, we present Quality-aware View Selection Lidar-Vision Odometry Risk Mapping (QVS-LIVO RiskMap), a multi-camera localization and dual-scale risk-mapping framework. It separates visual state-update selection from semantic observation utilization while sharing pose estimates and registered point clouds. The main contributions are as follows:

\begin{enumerate}
\setlength{\itemsep}{0.3em}

\item \textbf{Selective visual state updates via health- and complementarity-aware view selection}.
The front end performs view-specific, read-only pretracking across four non-overlapping cameras on a shared LIO posterior. Candidates are evaluated jointly by observation health and geometric complementarity to weakly constrained lidar pose directions. Four-state channel health management and switching hysteresis govern the selection of at most one eligible view for direct photometric correction. The correction is committed only after post-solve acceptance checks; the current LIO posterior is retained when no view is eligible or the correction is rejected.

\item \textbf{Dual-scale semantic risk mapping via available-view aggregation and geometry-constrained confirmation}.
Independently of localization-view selection, the mapping pipeline integrates semantic observations from all available views into a world-anchored 2.5D grid. Source-timestamp pairing, spatial projection and temporally decayed accumulation support this integration. Bounded semantic risk corrections preserve direct hard-obstacle constraints. Hard obstacles meeting the occupancy-confidence threshold are confirmed immediately, whereas slope- or step-induced candidates require additional spatial and temporal support. By preserving continuous risk, confirmation, and unknown states, the mapping pipeline produces a Local Risk BEV and a global 2.5D Semantic--Terrain Risk Map with Obstacle Instance Voxels.

\item \textbf{Systematic experimental validation via real-world field benchmarks, controlled camera faults, and onboard deployment}.
Sequences collected with the Multi-sensor AMR platform support internal mode comparisons and benchmarking against external SLAM baselines. Twelve controlled single-camera fault configurations, covering dropout, black frames and timestamp shifts, assess trajectory continuity and localization error. Onboard deployment on an NVIDIA Jetson AGX Orin benchmarks the generation latencies of both local and global maps, verifying the real-time performance of the complete system.

\end{enumerate}

The remainder of this paper is organized as follows. Section~\ref{sec:system-overview} introduces the platform and sensor configuration. Section~\ref{sec:methodology} presents the localization and risk-mapping methods. Section~\ref{sec:results-discussion} reports the experiments and discusses their implications and limitations. Section~\ref{sec:conclusion} concludes the paper.

\begin{figure}[!t]
\centering
\includegraphics[width=\columnwidth]{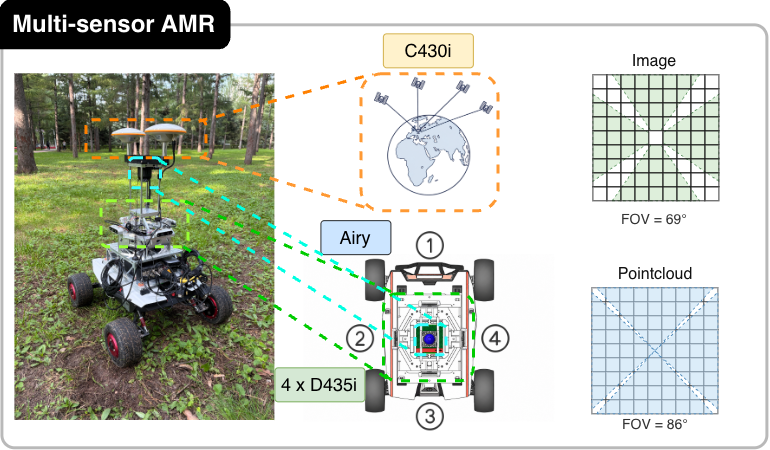}
\caption{Multi-sensor AMR platform and sensor coverage. Left: the robot in a woodland setting. Center: the C430i reference system, Airy lidar and four D435i cameras numbered front (1), left (2), rear (3) and right (4). Right: schematic horizontal coverage of RGB images (\(69^\circ\) per camera) and lidar point clouds (\(86^\circ\) per sector). Four support pillars create blind zones between the lidar coverage sectors.}
\label{fig:platform}
\end{figure}

\begin{figure*}[!t]
\centering
\includegraphics[width=\textwidth]{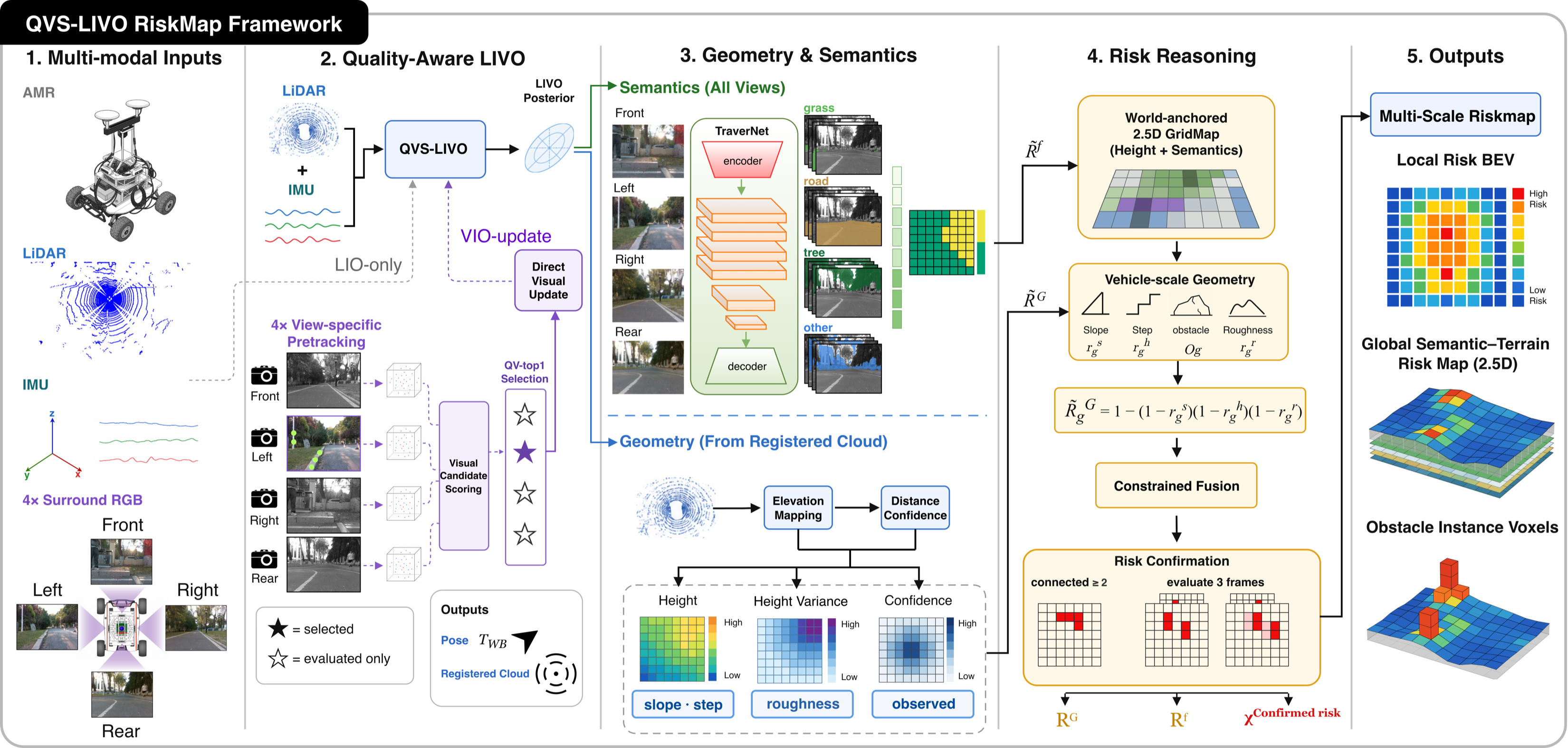}
\caption{QVS-LIVO RiskMap framework. Selective visual updates and available-view semantic mapping share poses and registered points. Vehicle-scale geometry constrains semantic risk fusion, followed by spatial and temporal confirmation. Outputs comprise Local Risk BEV, Global Semantic--Terrain Risk Map and Obstacle Instance Voxels.}
\label{fig:framework}
\end{figure*}

\section{System Overview}
\label{sec:system-overview}

In this paper, a multi-sensor Autonomous Mobile Robot (AMR) platform in Figure~\ref{fig:platform} is used for off-road data collection and onboard deployment. It is built on an AgileX Scout Mini chassis and carries an Airy lidar with an integrated IMU and four Intel RealSense D435i cameras. The cameras face front (1), left (2), rear (3) and right (4), following the numbering in the figure. QVS-LIVO uses point clouds and inertial measurements from the Airy, together with the camera RGB streams, for state estimation. The Airy's internal IMU supplies angular velocity and acceleration measurements, while its point clouds provide geometric constraints. The RGB streams provide photometric observations for visual updates and semantic observations for mapping.

Each RGB camera has a horizontal field of view of \(69^\circ\), producing non-overlapping views with gaps between adjacent sectors. In the installed configuration, four support pillars occlude the lidar, leaving four horizontal point-cloud coverage sectors of \(86^\circ\), separated by blind zones. Figure~\ref{fig:platform} illustrates the camera coverage and the lidar coverage remaining after structural occlusion.

The cameras acquire images independently, without inter-camera hardware synchronization. Source timestamps are used to associate images within a finite soft-synchronization window and align them with Airy lidar and IMU measurements for state estimation. An NVIDIA Jetson AGX Orin runs the QVS-LIVO front end and semantic--geometric risk-mapping pipeline onboard. Offline semantic replay uses an Intel Core i7-14700KF, an NVIDIA RTX 4090 with 24 GB of GPU memory and 64 GB of system memory.

An independent C430i RTK-GNSS/INS system (with its own internal IMU) logs reference positions at 1\,Hz for offline horizontal trajectory evaluation; QVS-LIVO and the mapping pipeline rely solely on the onboard Airy lidar/IMU and D435i cameras. Its two antennas are mounted above the other sensors.

\section{Methodology}
\label{sec:methodology}
\raggedbottom

The proposed QVS-LIVO RiskMap pipeline is presented in this section. Figure~\ref{fig:framework} shows how the pipeline extends the lidar-inertial estimation and direct photometric update framework of FAST-LIVO2. The inherited estimator provides IMU propagation, lidar updates and a shared LIO posterior. QVS-LIVO adds prior-preserving evaluation of multiple camera candidates, health- and complementarity-aware selection and conservative admission of at most one view to each visual correction. Independently, the mapping back end associates semantic observations from all available views with the resulting poses and registered point clouds, then applies vehicle-scale geometric analysis, constrained fusion and spatiotemporal risk confirmation. Sections~\ref{sec:qvslivo} and~\ref{sec:visual-selection} present view evaluation and accepted visual updates. Sections~\ref{sec:appearance}--\ref{sec:risk-confirmation} describe semantic accumulation and risk-map construction.

\subsection{Health- and Complementarity-Aware View Evaluation}
\label{sec:qvslivo}

\begin{figure*}[!t]
\centering
\includegraphics[width=\textwidth]{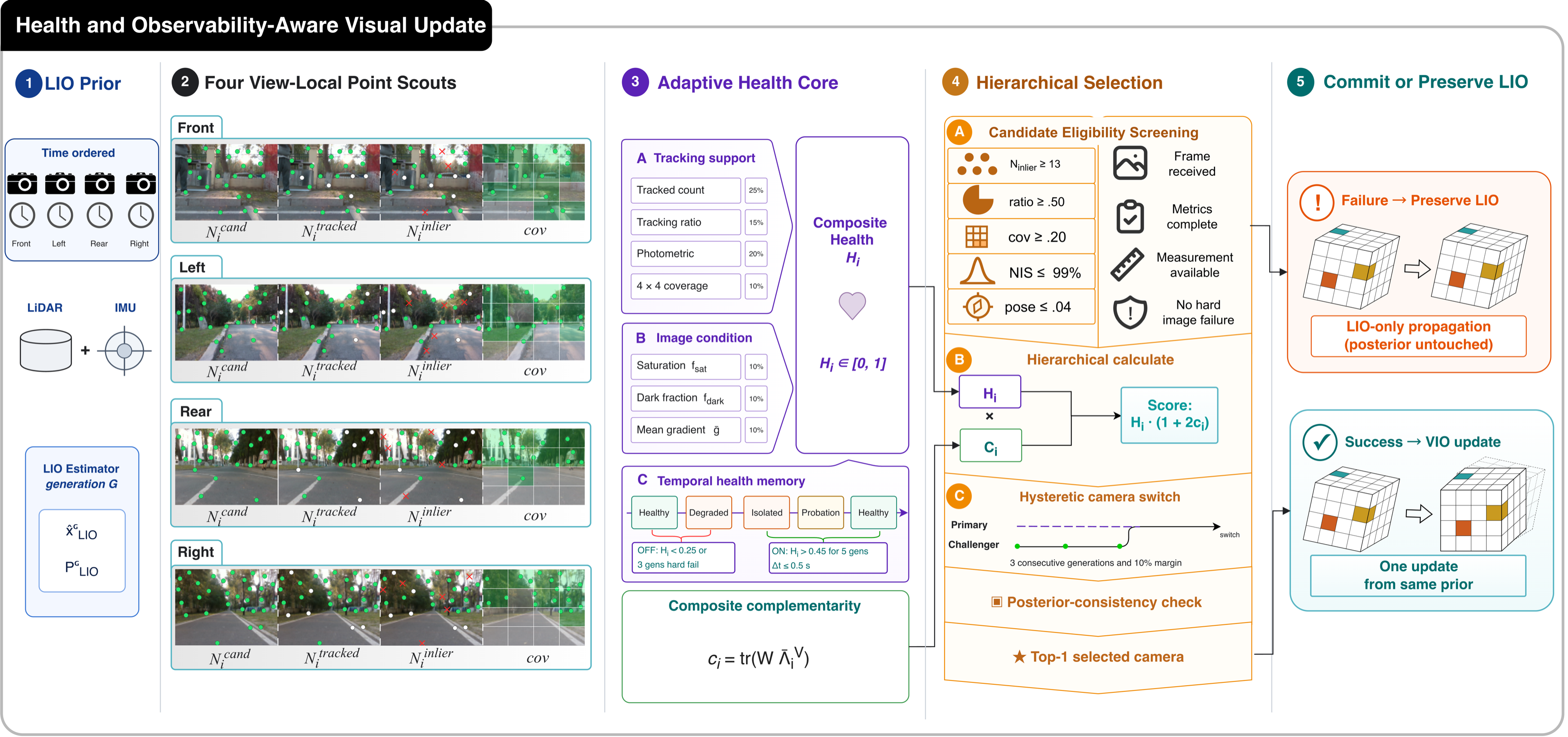}
\caption{QVS-LIVO visual-update decisions. View-specific pretracking supplies observation-health and weak-direction complementarity evidence on a shared LIO posterior. Channel reliability, candidate screening and hysteretic selection determine at most one visual correction. Posterior-consistency and post-solve checks determine whether the correction is committed.}
\label{fig:health-visual-update}
\end{figure*}

QVS-LIVO evaluates camera observations before admitting them to the visual state update. The evaluation combines observation health with geometric complementarity to weakly constrained lidar pose directions. As illustrated in Figure~\ref{fig:health-visual-update}, each available view undergoes prior-preserving pretracking on the same lidar-inertial posterior. The pretracking stage evaluates residuals and pose Jacobians without modifying the posterior or active visual map, thereby providing a common reference for candidate comparison.

The four cameras acquire images independently. Source-timestamp ordering groups at most one image per camera within \([t_0,t_0+2\Delta_s]\), where \(t_0\) is the earliest candidate timestamp and \(\Delta_s\) is the soft-synchronization tolerance. These images form the candidate batch \(\mathcal B_k\), whose timestamp is the midpoint of its earliest and latest observations. IMU propagation and lidar scan segmentation provide the corresponding posterior \(\mathcal X_k^L=(\mathbf x_k^L,\mathbf P_k^L)\), where \(k\) identifies the posterior generation. This posterior serves as the common prior for candidate evaluation and the subsequent visual correction.

View-specific pretracking records candidate points, successfully tracked points and robust inliers, denoted by \(N_i^{\mathrm{cand}}\), \(N_i^{\mathrm{tracked}}\) and \(N_i^{\mathrm{inlier}}\), respectively. It also evaluates the photometric residual RMSE \(e_i\), the overexposed- and dark-pixel fractions \(f_i^{\mathrm{sat}}\) and \(f_i^{\mathrm{dark}}\), the mean image gradient \(\bar g_i\), and the occupied fraction \(a_i\) of a \(4\times4\) image partition. The tracking ratio is \(\rho_i=N_i^{\mathrm{tracked}}/\max(N_i^{\mathrm{cand}},1)\). These measurements characterize the quantity, consistency, image condition and spatial distribution of the available visual evidence.

Each raw metric \(x_i^j\) is mapped to a health score \(s_i^j\in[0,1]\) using its reference value \(r_j\). Let \(\mathcal J_+=\{n,\rho,g,a\}\) collect the metrics for which larger values are healthier, and let \(\mathcal J_-=\{\mathrm{sat},\mathrm{dark}\}\) collect the two adverse image fractions. \(\operatorname{clip}(x)=\operatorname{clip}_{[0,1]}(x)\) is defined as \(\operatorname{clip}(x)=x\) if $x \in (0,1)$, and $0$ or $1$ if $x<0$ or $x>1$. Then the mappings are compactly written as
\begin{equation}
s_i^j=\begin{cases}
\operatorname{clip}(x_i^j/r_j),&j\in\mathcal J_+,\\
\operatorname{clip}(1-x_i^j/r_j),&j\in\mathcal J_-,\\
\left[1+(x_i^j/r_j)^2\right]^{-1},&j=e.
\end{cases}
\label{eq:health-metric-mapping}
\end{equation}
The raw metrics (cf., Table~\ref{tab:view-health-weights}) corresponding to \(j=\{n,\rho,e,\mathrm{sat},\mathrm{dark},g,a\}\) are \(\{N_i^{\mathrm{tracked}},\rho_i,e_i,f_i^{\mathrm{sat}},f_i^{\mathrm{dark}},\bar g_i,a_i\}\).  Larger scores consistently indicate healthier evidence. For camera \(i\), the available scores are aggregated as
\begin{equation}
h_i=\frac{\sum_{j\in\mathcal A_i}w_js_i^j}
{\sum_{j\in\mathcal A_i}w_j}.
\label{eq:observation-health}
\end{equation}
Here, \(\mathcal A_i\) contains the available metrics and \(w_j\) denotes their weights. Table~\ref{tab:view-health-weights} gives the seven aggregation weights. Table~\ref{tab:parameters} summarizes the corresponding reference values and other numerical settings used in the experiments. The denominator normalizes the available evidence; candidate eligibility additionally requires the checks described in Section~\ref{sec:visual-selection}.

\begin{table}[tb]
\centering
\caption{Observation-health weights used in view evaluation. Each metric is normalized before aggregation in (\ref{eq:observation-health}).}
\label{tab:view-health-weights}
\small
\setlength{\tabcolsep}{3pt}
\renewcommand{\arraystretch}{1.08}
\begin{tabular}{p{0.43\columnwidth}p{0.34\columnwidth}r}
\toprule
\textbf{Metric} & \textbf{Healthier evidence} & \textbf{Weight} \\
\midrule
Tracked point count & More tracked points & 25\% \\
Tracking ratio & Higher success ratio & 15\% \\
Photometric residual & Lower residual & 20\% \\
Spatial coverage & Wider image coverage & 10\% \\
Overexposed-pixel ratio & Less overexposure & 10\% \\
Dark-pixel ratio & Fewer dark pixels & 10\% \\
Mean gradient & Stronger texture & 10\% \\
\bottomrule
\end{tabular}
\end{table}

\begin{table*}[!t]
\centering
\caption{Implementation parameters used in the reported QVS-LIVO RiskMap experiments.}
\label{tab:parameters}
\scriptsize
\setlength{\tabcolsep}{5pt}
\renewcommand{\arraystretch}{1.10}
\begin{adjustbox}{max width=\textwidth}
\begin{tabular}{m{0.20\textwidth}m{0.34\textwidth}m{0.16\textwidth}m{0.20\textwidth}}
\toprule
\textbf{Stage} & \textbf{Parameter} & \textbf{Symbol} & \textbf{Value} \\
\midrule
Candidate formation & Soft-synchronization tolerance, reorder wait & $\Delta_s,\ \Delta_r$ & 20 ms, 20 ms \\
\cmidrule(lr){1-4}
\multirow[c]{2}{*}[-7pt]{Observation health} & Tracked-count, tracking-ratio and residual references & $N_0,\ \rho_0,\ e_0$ & 24, 0.50, 10 grayscale units \\
 & Overexposure, darkness, gradient and coverage references & $f_0^{\mathrm{sat}},\ f_0^{\mathrm{dark}},\ g_0,\ a_0$ & 0.25, 0.35, 12, 0.20 \\
\cmidrule(lr){1-4}
Information scaling & Rotation and translation characteristic scales & $s_R,\ s_t$ & $\pi/180$ rad, 0.1 m \\
\cmidrule(lr){1-4}
Candidate ranking & Complementarity weight, switching margin and persistence & $\gamma,\ m,\ N_s$ & 2, 0.10, 3 valid cycles \\
\cmidrule(lr){1-4}
\multirow[c]{2}{*}[-2pt]{Visual update} & Health floor, maximum covariance inflation & $h_{\min},\ \kappa_{\max}$ & 0.10, 10 \\
 & Pretracking point budget & -- & 240 total, max. 60/view \\
\cmidrule(lr){1-4}
\multirow[c]{2}{*}[-5pt]{Semantic association} & Pairing window, evidence half-life & $\Delta_p,\ \tau_{1/2}$ & 0.5 s, 5 s \\
 & Distance weighting & $\eta(d)$ & 1 to 3 m, linear to 0.25 at 12 m \\
\cmidrule(lr){1-4}
\multirow[c]{2}{*}[-2pt]{Geometric risk} & Slope and roughness saturation & $\theta_{\mathrm{crit}},\ v_{\mathrm{crit}}$ & $30^\circ$, $0.05\,\mathrm{m}^2$ \\
 & Step-risk transition, hard-obstacle height & $[h_{\mathrm{safe}},h_{\mathrm{crit}}]$ & 0.04--0.10 m, 0.25 m \\
\cmidrule(lr){1-4}
Risk fusion & Maximum semantic weight & $w_a$ & 0.35 \\
\cmidrule(lr){1-4}
Risk confirmation & Connected support, temporal votes & -- & 2 cells, 2 of 3 evaluations \\
\cmidrule(lr){1-4}
Map publication & Configured local and global rates & -- & Local $\sim$10 Hz, global $\sim$2 Hz \\
\bottomrule
\end{tabular}
\end{adjustbox}
\end{table*}

A healthy view may predominantly reinforce pose directions already constrained by lidar. QVS-LIVO therefore evaluates visual information along weak lidar directions. Let \(\mathbf J_{L,\ell}\) and \(\mathbf J_{V,i,p}\) denote the whitened six-dimensional pose-Jacobian blocks of lidar residual \(\ell\) and pretracking photometric residual \(p\), respectively. The corresponding pose information matrices are
\begin{equation}
\symbf{\Lambda}_k^L=\sum_{\ell\in\mathcal L_k}\mathbf J_{L,\ell}^{\mathsf T}\mathbf J_{L,\ell},
\qquad
\symbf{\Lambda}_{i,k}^V=\frac{1}{A_p}\sum_{p\in\mathcal P_{i,k}}w_{i,p}\mathbf J_{V,i,p}^{\mathsf T}\mathbf J_{V,i,p},
\label{eq:pose-information}
\end{equation}
where \(w_{i,p}\) is the robust residual weight and \(A_p\) normalizes the correlated pixels within an image patch. Suppressing the generation index below, define the dimensionless pose perturbation \(\delta\bar{\boldsymbol\xi}\) through \(\delta\boldsymbol\xi=\mathbf S\delta\bar{\boldsymbol\xi}\). Rotation--translation normalization then gives
\begin{equation}
\begin{aligned}
\mathbf S&=\operatorname{diag}(s_R\mathbf I_3,s_t\mathbf I_3),\\
\bar{\symbf{\Lambda}}^L&=\mathbf S^{\mathsf T}\symbf{\Lambda}^L\mathbf S,\\
\bar{\symbf{\Lambda}}_i^V&=\mathbf S^{\mathsf T}\symbf{\Lambda}_i^V\mathbf S,
\end{aligned}
\label{eq:information-scaling}
\end{equation}
where \(s_R\) and \(s_t\) are characteristic rotation and translation scales used to express the six pose perturbations in comparable normalized coordinates. Their experimental values are listed in Table~\ref{tab:parameters}. The normalized lidar information defines the weak-direction weighting matrix
\begin{equation}
\begin{aligned}
\bar{\symbf{\Lambda}}^L&=\mathbf U\operatorname{diag}(\lambda_j)\mathbf U^{\mathsf T},\\
\omega_j&=\operatorname{clip}
\left(\frac{\tau-\lambda_j}{\tau+\epsilon}\right),\\
\mathbf W&=\mathbf U\operatorname{diag}(\omega_j)\mathbf U^{\mathsf T}.
\end{aligned}
\label{eq:weak-directions}
\end{equation}
The eigenvalues \(\lambda_j\) and corresponding columns of \(\mathbf U\) describe the six pose-information directions. We use \(\tau=\max(0.05\lambda_{\max},10^{-9})\) and \(\epsilon=10^{-9}\). Directions below \(\tau\) receive larger weights as their lidar information decreases.

For the eligible camera set \(\mathcal E_k\), defined in Section~\ref{sec:visual-selection}, geometric complementarity, its within-batch normalization and the combined selection score are
\begin{equation}
\begin{aligned}
c_i&=\operatorname{tr}\!\left(\mathbf W\bar{\symbf{\Lambda}}_i^V\right),\\
\widetilde c_i&=\frac{c_i}{\max_{j\in\mathcal E_k}c_j+\epsilon},\\
q_i&=h_i\left(1+\gamma\widetilde c_i\right).
\end{aligned}
\label{eq:selection-score}
\end{equation}
The complementarity term favors views supporting weak lidar directions, while \(h_i\) retains observation reliability in the ranking. The normalized score \(\widetilde c_i\in[0,1]\) is relative to the currently eligible views, and \(\gamma\) controls its contribution. When complementarity scores vanish across all views, the ranking reduces to observation health, ensuring that view selection remains well-defined even under isotropic lidar constraints.

\subsection{Hierarchical View Selection and Accepted Visual Updates}
\label{sec:visual-selection}

QVS-LIVO separates candidate eligibility, view selection and update acceptance. A finite-state channel-reliability model and current observation checks determine which views may compete. Switching hysteresis then selects at most one view, and post-solve checks determine whether its correction is committed. Figure~\ref{fig:health-visual-update} summarizes this sequence, while Algorithm~\ref{alg:camera-selection} details camera selection.

Let \(z_{i,k}^{\mathrm{ch}}\) denote the reliability state of camera \(i\) at posterior generation \(k\). The model has four states: Healthy, Degraded, Isolated and Probation. State transitions depend on the current inspection and pretracking evidence \(\mathbf y_{i,k}\) together with the health score \(h_i\). Persistent low health or unsuccessful tracking identifies sustained degradation, whereas hard image faults trigger isolation. A recovering channel enters Probation and regains eligibility only after satisfying the required healthy-observation count and recovery duration. Interrupted recovery returns it to Isolated, preventing transient recovery from immediately restoring a failed channel to visual estimation.

Current candidates are screened for temporal association, image validity, tracking support, residual consistency and information-matrix validity. Their reliability states must also permit visual use. Candidates satisfying these conditions form \(\mathcal E_k\), over which (\ref{eq:selection-score}) is evaluated. The finite-state model governs eligibility over time, switching hysteresis governs competition among eligible views.

Let \(i_{k-1}^*\) denote the previously selected camera. If it remains eligible and challengers exist, the strongest challenger is \(j_k=\arg\max_{i\in\mathcal E_k\setminus\{i_{k-1}^*\}}q_i\). A switch requires \(q_{j_k}>(1+m)q_{i_{k-1}^*}\) over \(N_s\) consecutive valid selection cycles, where \(m\) is the score margin and \(N_s\) is the persistence length. The selected view is
\begin{equation}
i_k^*=\begin{cases}
\varnothing,&\mathcal E_k=\varnothing,\\
\arg\max_{i\in\mathcal E_k}q_i,
&\mathcal E_k\ne\varnothing,\ i_{k-1}^*\notin\mathcal E_k,\\
j_k,&\text{the persistence condition holds},\\
i_{k-1}^*,&\text{otherwise}.
\end{cases}
\label{eq:top1-selection}
\end{equation}
This rule retains an eligible view through short-lived score fluctuations and permits immediate reselection when that view becomes ineligible. An empty candidate set produces no visual solve.

To prevent an asynchronously completed visual correction from being applied to an outdated prior, the selected candidate is admitted only when its batch identity, camera identity, image timestamps and LIO posterior generation remain consistent with those used during evaluation. A mismatch rejects the visual correction and preserves the current LIO posterior.

For the selected camera \(j=i_k^*\), observation health also determines the visual covariance inflation factor,
\begin{equation}
\kappa_j=\min\!\left(\kappa_{\max},\frac{1}{\max(h_j,h_{\min})}\right),
\label{eq:health-inflation}
\end{equation}
Using \(\mathbf P_L=\mathbf P_k^L\), the direct photometric correction solves
\begin{equation}
\begin{aligned}
\delta\mathbf x^*&=\arg\min_{\delta\mathbf x}\Big\{
\|\delta\mathbf x-\delta\mathbf x_0\|_{\mathbf P_L^{-1}}^2+\sum_{p\in\mathcal P_j}\rho\!\left(
\frac{r_p(\delta\mathbf x)^2}{A_p\kappa_j}\right)\Big\}.
\end{aligned}
\label{eq:single-winner-update}
\end{equation}
Here, \(\delta\mathbf x\) is the error-state correction, and \(\delta\mathbf x_0\) represents the current linearization error relative to the LIO prior. The set \(\mathcal P_j\) contains valid photometric residuals from the selected view. Each \(r_p\) compares the current image with its reference patch, \(A_p\) provides noise normalization, and \(\rho\) is the robust loss. Lower health increases \(\kappa_j\), reducing visual influence while retaining the LIO posterior as the prior constraint.

After solving, acceptance checks evaluate residuals, pose increments and the numerical validity of the normal equations, and reverify posterior consistency. An accepted correction commits the state and the corresponding visual-map update. Otherwise, the system retains \((\mathbf x_k^L,\mathbf P_k^L)\) and discards the proposed correction. Subsequent IMU propagation, lidar updates and point-cloud registration continue.

\begin{algorithm}[tb]
\caption{Health- and Complementarity-Aware Camera Selection}
\label{alg:camera-selection}
\begingroup
\algrenewcommand\algorithmicrequire{\textbf{Input:}}
\algrenewcommand\algorithmicensure{\textbf{Output:}}
\begin{algorithmic}[1]
\Require Image batch $\mathcal B_k$, shared LIO posterior
\Statex $\mathcal X_k^L=(\mathbf x_k^L,\mathbf P_k^L)$, lidar information $\boldsymbol\Lambda_k^L$,
\Statex channel states $\mathbf z_k^{\mathrm{ch}}$, selection memory $\mathcal M_k$
\Ensure Selected camera $i_k^*$ or $\varnothing$, updated $\mathbf z_k^{\mathrm{ch}},\mathcal M_k$
\State $\mathcal E_k\gets\varnothing$, $q\gets\varnothing$
\State $\mathbf W_k^L\gets\Call{WeakDirections}{\boldsymbol\Lambda_k^L}$
\For{each camera $i$ represented in $\mathcal B_k$}
    \State $\mathbf y_i\gets\Call{InspectAndPretrack}{\mathcal B_k[i],\mathcal X_k^L}$
    \State $h_i\gets\Call{ObservationHealth}{\mathbf y_i}$
    \State $z_{i,k}^{\mathrm{ch}}\gets\Call{UpdateChannelState}{z_{i,k}^{\mathrm{ch}},\mathbf y_i,h_i}$
    \If{$\Call{Eligible}{\mathbf y_i,z_{i,k}^{\mathrm{ch}}}$}
        \State $c_i\gets\Call{Complementarity}{\mathbf y_i,\mathbf W_k^L}$
        \State $\mathcal E_k\gets\mathcal E_k\cup\{i\}$
    \EndIf
\EndFor
\If{$\mathcal E_k\ne\varnothing$}
    \For{each $i\in\mathcal E_k$}
        \State $\widetilde c_i\gets c_i/(\max_{j\in\mathcal E_k}c_j+\epsilon)$
        \State $q_i\gets h_i(1+\gamma\widetilde c_i)$
    \EndFor
\EndIf
\State $(i_k^*,\mathcal M_k)\gets\Call{HystereticSelect}{\mathcal E_k,q,\mathcal M_k}$
\State \Return $i_k^*,\mathbf z_k^{\mathrm{ch}},\mathcal M_k$
\end{algorithmic}
\endgroup
\end{algorithm}

Algorithm~\ref{alg:camera-selection} uses \(\mathbf z_k^{\mathrm{ch}}\) for channel-reliability states and \(\mathcal M_k\) for the previous selection and hysteresis memory. The evidence vector \(\mathbf y_i\) is obtained by inspection and pretracking on \(\mathcal X_k^L\). Its returned view identifies the candidate for the visual solve, the posterior-consistency and acceptance checks above determine whether that solve changes the state. Unselected views supply only evaluation evidence within visual estimation, while remaining available to the semantic mapping pipeline.

\subsection{Available-View Semantic Evidence Accumulation}
\label{sec:appearance}
\label{sec:semantic-representation}

The mapping branch uses semantic observations from all available cameras, independently of localization-view selection. TraverNet (see Figure~\ref{fig:framework}, `3. Geometry \& Semantics') adapts the Ultralytics YOLO26s dense semantic model to four classes: grass, road, tree and other~\citep{Ultralytics2026Semantic}. Each image produces an original-resolution class mask \(M_i(u,v)\) and confidence map \(Q_i(u,v)\). Here, \(Q_i(u,v)\) denotes the maximum class probability, which serves as a relative evidence weight during temporal accumulation. %Invalid or ambiguous labels use \texttt{ignore} (255), which is not a fifth predicted class.
Invalid or ambiguous labels are discarded in the experiments.
The detailed implementation settings are summarized in Section~\ref{sec:runtime-measurements}.

Each class mask and confidence map retains the source timestamp, camera identity and original image dimensions of its input. Spatial association therefore uses acquisition time rather than inference-completion time.

Following the shared-grid organization of multi-view representations~\citep{Li2024FastBEV}, semantic evidence is associated with world-coordinate registered points. For a point cloud at \(t_p\), each camera supplies its nearest semantic result within the pairing window \(\Delta_p\), class and confidence timestamps must agree. Projection uses the pose at the semantic-frame time and the camera calibration. A world point \(\mathbf p\) projects to pixel \((u_i,v_i)\) at positive optical depth \(d_i\), yielding distance-weighted confidence \(\bar Q(\mathbf p)=Q_i(u_i,v_i)\eta(d_i)\). The bounded weight \(\eta(d_i)\) retains near-range evidence and attenuates distant observations according to the breakpoints in Table~\ref{tab:parameters}. If multiple valid views cover a point, the view with the smallest optical depth is used. Previously accumulated ground points are also projected to extend semantic coverage of existing geometry.

Within each world-grid cell \(g\), weighted class votes and effective point-observation counts decay with the same half-life. Superscripts \(-\) and \(+\) denote pre-update and updated values. With \(\beta=2^{-\Delta t/\tau_{1/2}}\), where \(\tau_{1/2}\) is the semantic-evidence half-life and \(\Delta t\) is the elapsed update time, accumulation and readout are
\begin{equation}
\begin{aligned}
v_{g,c}^+&=\beta v_{g,c}^-+\sum_{j\in\mathcal O_g}\bar Q_j\mathbf1[c_j=c],\\
n_g^+&=\beta n_g^-+|\mathcal O_g|,\\
c_g^S&=\arg\max_c v_{g,c}^+,\qquad
q_g^S=\frac{v_{g,c_g^S}^+}{n_g^++\epsilon}.
\end{aligned}
\label{eq:semantic-accumulation}
\end{equation}
Here, \(\mathcal O_g\) contains the current point observations assigned to cell \(g\), each with class \(c_j\) and weighted confidence \(\bar Q_j\). The indicator \(\mathbf1[\cdot]\) is one when its condition holds. The denominator counts decayed point observations rather than summed class weights, so \(q_g^S\) reflects class agreement and per-observation confidence. Readout requires sufficient effective evidence, absent or sufficiently decayed evidence yields an unknown semantic state.

\subsection{Geometry-Constrained Semantic Risk Fusion}
\label{sec:geometry}
\label{sec:risk-mapping}

Registered point clouds supply geometry to the world-anchored 2.5D grid shown in Figure~\ref{fig:framework}. Each cell stores ground elevation, spatial height variance, observation confidence, above-ground structure and the semantic attributes from Section~\ref{sec:appearance}. Above-ground structure determines occupancy within the vehicle's passage band, while observation confidence distinguishes observed from unknown terrain.

Let \(z_g\) denote the ground elevation of cell \(g\), and let \(\mathcal N_h(g)\) and \(\mathcal Z_g\) denote its valid step-analysis neighborhood and local elevation samples, respectively. Vehicle-scale slope \(\theta_g\), height discontinuity \(\delta h_g\), and spatial roughness \(\sigma_{z,g}^2\) are computed as
\begin{equation}
\begin{aligned}
\theta_g&=\arctan\!\left(\lVert\nabla z_g\rVert_2\right),\\
\delta h_g&=\max_{j\in\mathcal N_h(g)}|z_g-z_j|,\\
\bar z_g&=\frac{1}{|\mathcal Z_g|}\sum_{z_j\in\mathcal Z_g}z_j,
\qquad
\sigma_{z,g}^2=\frac{1}{|\mathcal Z_g|}\sum_{z_j\in\mathcal Z_g}(z_j-\bar z_g)^2.
\end{aligned}
\label{eq:geometric-features}
\end{equation}
The gradient is evaluated by centered differences over the grid resolution, and roughness uses the local spatial elevation variance rather than the sensor's measurement-uncertainty layer. The corresponding normalized risk components are
\begin{equation}
\begin{aligned}
r_g^s&=\operatorname{clip}\!\left(\frac{\theta_g}{\theta_{\mathrm{crit}}}\right),\\
r_g^h&=\operatorname{clip}\!\left(\frac{\delta h_g-h_{\mathrm{safe}}}{h_{\mathrm{crit}}-h_{\mathrm{safe}}}\right),\\
r_g^r&=\operatorname{clip}\!\left(\frac{\sigma_{z,g}^2}{v_{\mathrm{crit}}}\right).
\end{aligned}
\label{eq:geometric-component-risks}
\end{equation}
Here, \(\theta_{\mathrm{crit}}\) and \(v_{\mathrm{crit}}\) are the slope and roughness saturation levels, while \([h_{\mathrm{safe}},h_{\mathrm{crit}}]\) defines the step-risk transition from the safe deadband to saturation. Cells without sufficient valid geometry retain an unknown component rather than being assigned zero risk. These components form a continuous geometric risk, with occupancy imposed separately:
\begin{equation}
\begin{aligned}
\widetilde R_g^G&=1-(1-r_g^s)(1-r_g^h)(1-r_g^r),\\
R_g^G&=O_g+(1-O_g)\widetilde R_g^G.
\end{aligned}
\label{eq:geometry-risk}
\end{equation}
The first line combines the three continuous components, as shown in Figure~\ref{fig:framework}. The binary occupancy flag \(O_g\) is one when above-ground structure intrudes into the passage band. Occupied cells therefore retain maximum geometric risk, independently of their semantic class. Step risk \(r_g^h\) and occupancy \(O_g\) remain distinct inputs.

Semantic observations provide a bounded correction to the geometric estimate. Class reliability and accumulated confidence determine the effective appearance weight:
\begin{equation}
\begin{aligned}
\alpha_g&=w_a\kappa^A(c_g^S)q_g^S,\\
R_g^F&=\begin{cases}
1,&O_g=1,\\
(1-\alpha_g)R_g^G+\alpha_gR^A(c_g^S),&O_g=0.
\end{cases}
\end{aligned}
\label{eq:constrained-fusion}
\end{equation}
Here, \(R^A(c)\) is the class-dependent appearance-risk prior and \(\kappa^A(c)\in[0,1]\) its reliability coefficient. The configured upper bound on \(w_a\), listed in Table~\ref{tab:parameters}, ensures that geometry retains the larger continuous weight, while weaker semantic evidence produces a smaller correction. Semantic classes neither reduce occupied-cell risk nor independently create direct hard obstacles. Continuous risk \(R_g^F\) and its confirmation status are retained separately.

\subsection{Spatiotemporal Risk Confirmation}
\label{sec:risk-confirmation}

Risk confirmation evaluates whether geometric evidence has sufficient support to establish high risk. Direct hard obstacles follow a separate path from slope- or step-induced candidates. For grid cell \(g\), \(t\) indexes successive evaluations of its grid state rather than individual camera or lidar frames. The hard-obstacle flag \(H_{g,t}\) is one when passage-band occupancy is present and its observation confidence \(q_{g,t}^{O}\) reaches the direct-evidence threshold \(q_0\). These obstacles are confirmed immediately.

Other candidates require connected spatial support and repeated votes within a recent evaluation window. Let \(r_{g,t}=\max(r_{g,t}^s,r_{g,t}^h)\), and let \(\theta\) denote the candidate risk threshold. The indicators \(D_{g,t}\), \(A_{g,t}\) and \(V_{g,t}\) represent valid direct geometric evidence, sufficient connected support and a candidate veto, respectively. The veto is supplied by high-confidence low-risk semantics or previously traversed footprints. Candidate votes and confirmation are
\begin{equation}
\begin{aligned}
C_{g,t}&=D_{g,t}A_{g,t}(1-V_{g,t})\mathbf1[r_{g,t}\geq\theta],\\
\chi_{g,t}&=H_{g,t}\lor\mathbf1[C_{g,t}+C_{g,t-1}+C_{g,t-2}\geq2].
\end{aligned}
\label{eq:risk-confirmation}
\end{equation}
Thus, \(C_{g,t}\) records the current candidate decision, while \(\chi_{g,t}\) combines direct hard-obstacle evidence with recent candidate votes. In the evaluated configuration, sufficient support means at least two connected cells and confirmation requires at least two positive votes among the three latest evaluations. Semantic evidence can suppress a non-hard-obstacle candidate vote but cannot revoke a direct hard obstacle or independently generate confirmed high risk. The temporal window in Figure~\ref{fig:framework} refers to grid evaluations. Cells without valid geometric observations retain an unknown map state, distinct from either confirmed or unconfirmed risk.

The common grid supports local queries and global accumulation. The robot-centered Local Risk BEV preserves confirmed high risk through max pooling and aggregates semantics through confidence-weighted voting. The Global Semantic--Terrain Risk Map stores ground, above-ground structure, semantics, occupancy and risk in world coordinates. Obstacle Instance Voxels retain the three-dimensional structure of obstacles. These outputs share poses and registered points, while local publication and global accumulation are scheduled separately.

\flushbottom

\subsection{Declaration of Generative AI Use}
\label{sec:ai-use}

ChatGPT (OpenAI) and DeepSeek were used to assist with code modification, data analysis, and English-language editing during the research and preparation of this manuscript. The authors reviewed and revised the AI-assisted outputs and take full responsibility for the final manuscript.

\section{Results and Discussion}
\label{sec:results-discussion}

This section evaluates the proposed framework on two field sequences. We assess horizontal localization accuracy and point-cloud consistency, evaluate trajectory continuity and localization error under controlled single-camera faults, and examine the dual-scale semantic–geometric risk-mapping outputs. Runtime performance and map publication rates are reported for workstation replay and onboard deployment on an AGX Orin. The section concludes with a discussion of the architectural implications and operational trade-offs.

\subsection{Experimental Setup}
\label{sec:experimental-setup}

\subsubsection{Platform, Datasets and Reference}
\label{sec:platform-datasets}

\begin{figure*}[!t]
\centering
\includegraphics[width=\textwidth]{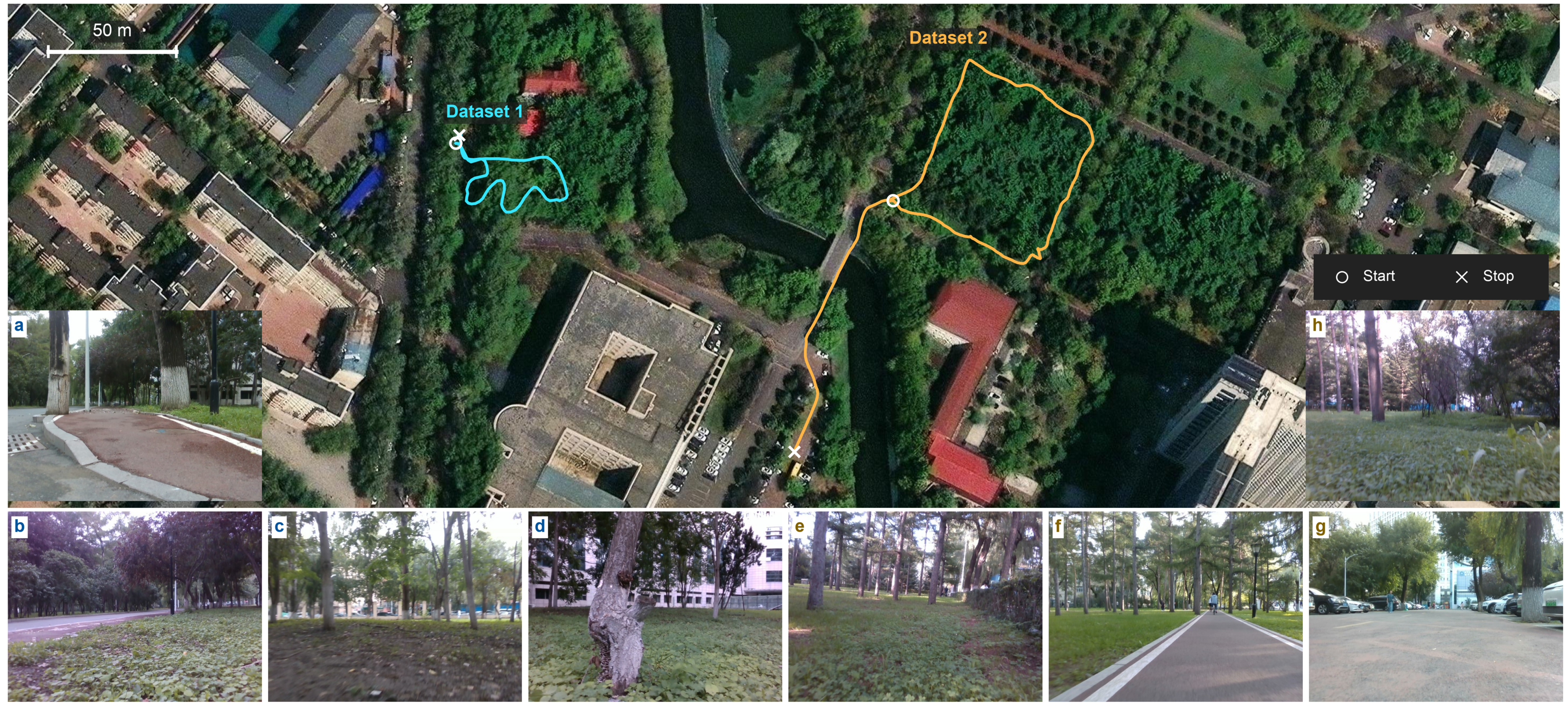}
\caption{Field acquisition areas and representative scenes from two field sequences containing both off-road terrain and paved surfaces. Blue and orange trajectories indicate the schematic acquisition routes for Dataset\#1 and Dataset\#2, respectively, circles and crosses mark the start and end points. Four camera observations from Dataset\#1 show (a) a paved intersection and curb, (b) leaf-covered ground beside a road, (c) tree trunks under the canopy with alternating bright and shaded areas, and (d) nearby trunks and low vegetation. Four observations from Dataset\#2 show (e) a woodland passage beside a hedge, (f) a paved tree-lined footpath, (g) a campus road with parked vehicles on both sides, and (h) grass-covered ground beneath trees. The scale bar applies to the satellite image.}
\label{fig:realworld}
\end{figure*}

Figure~\ref{fig:realworld} shows the acquisition areas, schematic routes and representative scenes for the two field sequences. The blue and orange trajectories correspond to Dataset\#1 and Dataset\#2, respectively.

Field data were collected using the multi-sensor AMR platform described in Section~\ref{sec:system-overview}, with an Airy lidar and four cameras facing front, left, rear and right. Dataset\#1 (178.78\,s) and Dataset\#2 (350.12\,s) contain both off-road terrain and paved surfaces, as illustrated in Figure~\ref{fig:realworld}. Trajectories for horizontal RMSE evaluation were provided by the independent 1\,Hz C430i RTK-GNSS/INS reference system.

\subsubsection{Comparison Protocols and Metrics}
\label{sec:comparison-protocols}

Four configurations based on FAST-LIVO2~\citep{Zheng2025FastLIVO2} were evaluated to compare different strategies for visual observation utilization:
\begin{enumerate}
\item \textbf{LIO}: Disables visual updates and retains the original lidar--inertial pipeline.
\item \textbf{Front VIO}: Uses only the forward-facing camera, corresponding to the original FAST-LIVO2 configuration.
\item \textbf{Joint VIO}: Jointly uses all four camera streams for visual estimation.
\item \textbf{QVS-LIVO}: Selects at most one eligible view based on observation health and geometric complementarity.
\end{enumerate}
The labels ``Front'' and ``Adaptive'' in the figures refer to Front VIO and QVS-LIVO, respectively.

External comparisons include ORB-SLAM2~\citep{MurArtal2017ORBSLAM2}, ORB-SLAM3~\citep{Campos2021ORBSLAM3}, LOAM~\citep{Zhang2014LOAM} and FAST-LIO2~\citep{Xu2022FastLIO2}. QVS-LIVO and the four external baselines were evaluated on the complete sequences of Dataset\#1 and Dataset\#2 under the current sensor configuration.

Localization accuracy was evaluated using horizontal trajectory RMSE relative to the C430i reference positions. Point-cloud consistency was assessed using local plane thickness P90, the 90th percentile of thickness, with a common region of interest (ROI) and voxel configuration. This metric characterizes local surface thickness and inter-frame ghosting.

\subsubsection{Controlled Camera Faults}
\label{sec:controlled-faults}

To distinguish the effects of input interruption, invalid image content and abnormal temporal association on localization, we introduced three controlled single-camera faults:
\begin{enumerate}
\item \textbf{RGB Dropout}: Simulates camera-stream or transmission interruption by withholding image message arrivals.
\item \textbf{Black Frames}: Retains image-message arrivals but replaces image content with black frames, approximating severe occlusion or exposure failure.
\item \textbf{Timestamp Shift ($+60$\,ms)}: Retains the original image content while adding a $+60$\,ms temporal offset to image timestamps, perturbing temporal correspondence with lidar and IMU measurements.
\end{enumerate}

The fault experiment was conducted on Dataset\#2. Within the source-time interval $[105, 120)$\,s, one fault was applied to one of the front, left, rear or right cameras at a time, producing 12 configurations; the original stream was restored outside that interval. Depth, lidar, IMU and the other camera inputs were left unchanged, controlling the source of the anomaly and testing whether remaining observations could sustain localization. The setting addresses short-term anomalies in a single RGB stream. % and does not cover simultaneous failures of multiple cameras or other sensors.

ORB-SLAM2 and ORB-SLAM3 used the affected camera's RGB-D input throughout each run, with normal-input controls for all four views to distinguish view-dependent effects from fault effects. ARTI-VIO~\citep{Feng2026ARTIVIO} was adapted to four RGB-D streams plus IMU. Fixed-front-view FAST-LIVO2 is reported only for front-camera faults, while camera-independent FAST-LIO2 uses its normal-run result as a reference. Beyond trajectory RMSE, failure modes including persistent tracking loss, processing backlog leading to trajectory termination, and invalid trajectory timestamps were recorded separately.

\subsubsection{Semantic Processing and Runtime Measurements}
\label{sec:runtime-measurements}

TraverNet was initialized from a Cityscapes-pretrained YOLO26s semantic model~\citep{Cordts2016Cityscapes,Ultralytics2026Semantic} and adapted using remapped annotations from RUGD, RELLIS-3D, GO, WildScenes, Freiburg Forest and YCOR~\citep{Wigness2019RUGD,Jiang2021RELLIS,Jiang2026GO,Vidanapathirana2025WildScenes,Valada2017Freiburg,Maturana2018YCOR}. The 48 locally labeled images were divided by scene--camera group into 37 training and 11 validation images. Off-road adaptation was followed by joint fine-tuning, the final stage used AdamW for three epochs, a learning rate of \(10^{-5}\), batch size 32 and input-size setting 640. The training list contained 4,596 sampled entries (4,337 unique paths), including eightfold sampling of the local training images, and the validation list contained 1,222 images.

Online semantic inference used one GPU-resident model with round-robin service of nonempty camera slots. In the latest-frame configuration, a newly received image replaced an older unprocessed image from the same camera, batch-enabled configurations used bounded per-camera queues. Semantic and BEV replay were evaluated on an Intel Core i7-14700KF with an NVIDIA RTX 4090 (24\,GB GPU memory) and 64\,GB system memory, while onboard deployment used an AGX Orin. Standalone inference time, full-stack semantic output rate, semantic projection output rate and timestamp-pairing rate were measured during replay. Map timing measurements and publication rates were also recorded, with the measurement scope specified for each configuration.

\subsection{Localization and Point-Cloud Consistency}
\label{sec:localization-results}

\subsubsection{Internal Mode Comparison}
\label{sec:internal-results}

\begin{figure}[!t]
\centering
\includegraphics[width=\columnwidth]{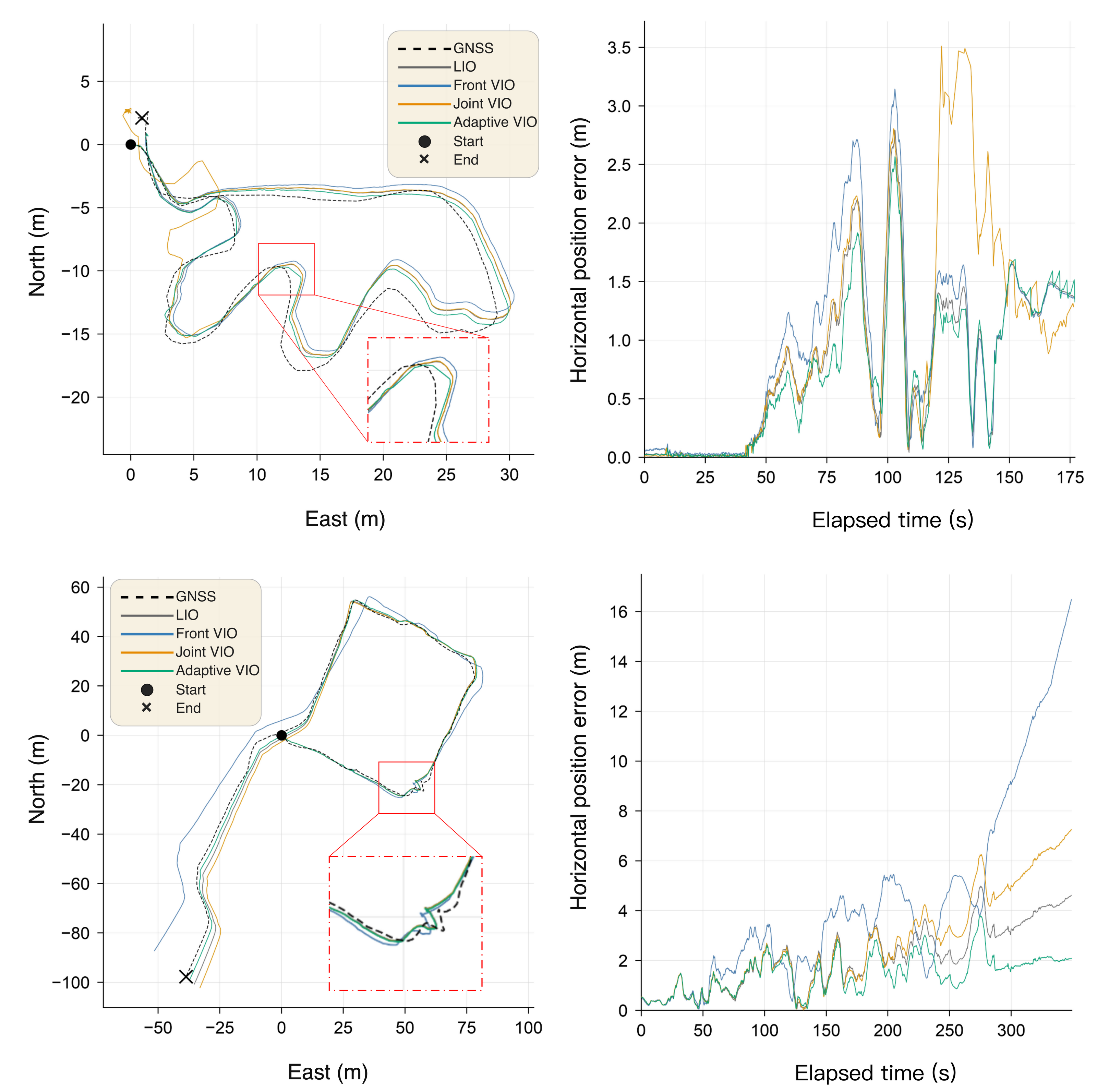}
\caption{Internal mode comparison on Dataset\#1 (top) and Dataset\#2 (bottom): trajectories with local zooms (left) and horizontal position errors over time (right). RMSE values are listed in Table~\ref{tab:gnss-results}. Adaptive VIO denotes QVS-LIVO.}
\label{fig:internal-mode-comparison}
\end{figure}

Table~\ref{tab:gnss-results} summarizes the internal comparison of four visual observation strategies on Dataset\#1 and Dataset\#2. QVS-LIVO achieved the lowest horizontal RMSE on both sequences, at 0.910\,m and 1.664\,m, respectively. Figure~\ref{fig:internal-mode-comparison} shows the corresponding estimated trajectories and horizontal position error profiles.

On Dataset\#1, all four configurations achieved horizontal RMSEs below 1.3\,m. Front VIO and Joint VIO had slightly higher errors than LIO, whereas QVS-LIVO achieved the lowest error. On Dataset\#2, Joint VIO achieved an RMSE of 3.306\,m, compared with 5.833\,m for Front VIO, but remained less accurate than LIO and QVS-LIVO. Joint fusion therefore did not consistently translate additional visual observations into lower trajectory error. Its vulnerability to localized channel failures is further examined in Section~\ref{sec:fault-results}, where single-camera dropout caused processing backlog and trajectory termination in the evaluated Joint VIO implementation. This observation supports the importance of handling channel failures when combining multiple visual streams.

To complement trajectory error, we estimated the effective visual-tracking time fraction, $\eta_{\mathrm V}$, from full-sequence tracking diagnostics. A record was valid when it reported at least one tracked point and a finite photometric residual. The analysis used front-view records for Front VIO, the union of valid per-camera records for Joint VIO, and selected-view records for QVS-LIVO. Each recorded state was held until the next evaluation. Valid interval durations were summed and divided by the full recorded sequence duration, without extending validity before the first or after the last diagnostic record. This metric quantifies the temporal continuity of valid visual tracking across the sequence.

The tracking fractions on Dataset\#1 were 44.9\%, 90.0\% and 86.6\% for Front VIO, Joint VIO and QVS-LIVO, respectively. The corresponding values on Dataset\#2 were 64.9\%, 96.0\% and 95.3\%. Joint VIO achieved the highest availability under the at-least-one-camera criterion, while QVS-LIVO maintained comparable availability through its selected view and achieved lower horizontal RMSE. These results distinguish the availability of visual tracking from the accuracy of the resulting trajectory estimate.

Colored-map output was evaluated by counting valid colored points after spatial voxel deduplication at the end of each full-sequence mapping replay. Let $\mathcal P_{\mathrm{RGB}}$ contain points with successfully assigned RGB colors and finite coordinates. At a common voxel resolution of $r=0.1\,\mathrm m$, the deduplicated count is
\begin{equation}
N_{\mathrm{color}}=\left|\left\{\left\lfloor\frac{\mathbf p}{r}\right\rfloor:\mathbf p\in\mathcal P_{\mathrm{RGB}}\right\}\right|.
\label{eq:colored-points}
\end{equation}
Here, $\mathbf p$ is a point in world coordinates, the floor operator acts componentwise, and $|\cdot|$ denotes set cardinality. Thus, $N_{\mathrm{color}}$ counts unique voxels containing valid colored points at the specified resolution. Valid black pixels were retained. Counts used the complete RGB point cloud without frame subsampling or per-frame point limits, with deduplication based on 64-bit three-dimensional voxel coordinates. The mapping replays used the same software version and input data across configurations. Front VIO used front-camera coloring, whereas Joint VIO and QVS-LIVO used four-view coloring. Colored-point counts came from these mapping replays, RMSEs came from the localization evaluations. LIO was excluded from the coloring comparison.

\begin{table}[!t]
\centering
\caption{Localization accuracy, colored-map output and visual-tracking availability.}
\label{tab:gnss-results}
\scriptsize
\setlength{\tabcolsep}{3.5pt}
\renewcommand{\arraystretch}{1.12}
\begin{adjustbox}{max width=\columnwidth}
\begin{tabular}{lrrrrrr}
\toprule
\textbf{Method} & \multicolumn{3}{c}{\textbf{Dataset\#1}} & \multicolumn{3}{c}{\textbf{Dataset\#2}} \\
 & RMSE & $N_{\mathrm{color}}$ & $\eta_{\mathrm V}$ & RMSE & $N_{\mathrm{color}}$ & $\eta_{\mathrm V}$ \\
\midrule
LIO & 1.042 & -- & -- & 2.465 & -- & -- \\
Front VIO~\citep{Zheng2025FastLIVO2} & 1.222 & 259,032 & 44.9 & 5.833 & 736,335 & 64.9 \\
Joint VIO & 1.190 & 603,894 & \textbf{90.0} & 3.306 & 2,291,670 & \textbf{96.0} \\
QVS-LIVO & \textbf{0.910} & 579,315 & 86.6 & \textbf{1.664} & 2,221,868 & 95.3 \\
\bottomrule
\end{tabular}
\end{adjustbox}
\par\vspace{3pt}
\begin{minipage}{\columnwidth}
\footnotesize
RMSE is in meters, $N_{\mathrm{color}}$ counts unique colored voxels at 0.1 m resolution, $\eta_{\mathrm V}$ is the tracking-time fraction (\%). Tracking uses the front view (Front VIO), any valid view (Joint VIO), or the selected view (QVS-LIVO). --: not applicable. Bold: lowest RMSE or highest $\eta_{\mathrm V}$.
\end{minipage}
\end{table}

QVS-LIVO retained 2.24 and 3.02 times as many valid colored points as Front VIO on Dataset\#1 and Dataset\#2, respectively. These correspond to increases of approximately 123.6\% and 201.7\%. Joint VIO also used four-view coloring and retained slightly more points than QVS-LIVO. These results confirm that decoupling mapping from single-view localization allows QVS-LIVO to preserve surround-view visual coverage comparable to Joint VIO while maintaining superior pose accuracy.

\begin{figure*}[!t]
\centering
\includegraphics[width=\textwidth]{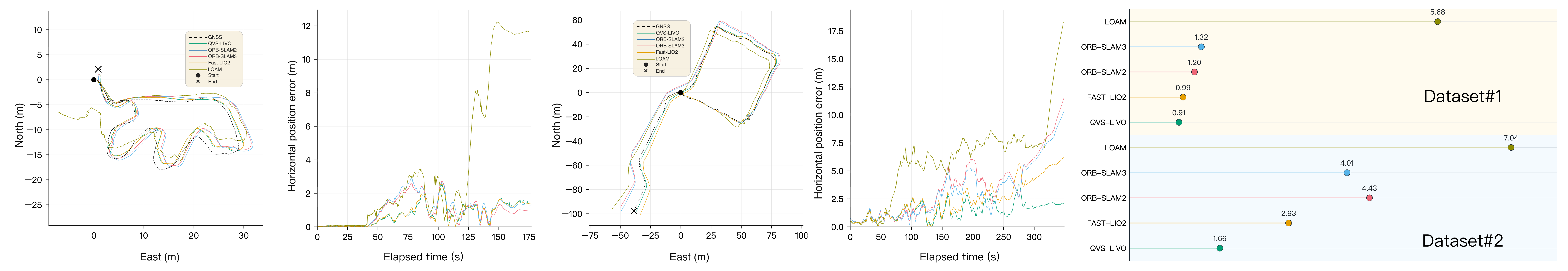}
\caption{Comparison with four external baselines on Dataset\#1 and Dataset\#2. Trajectories and horizontal position errors over time are shown on the left, and RMSE on the right. FAST-LIO2 uses the vehicle-compatible variant, all methods share the reference and alignment rule.}
\label{fig:external-baselines}
\end{figure*}

\subsubsection{Comparison with External SLAM Baselines}
\label{sec:external-results}

Figure~\ref{fig:external-baselines} compares QVS-LIVO with ORB-SLAM2, ORB-SLAM3, LOAM and FAST-LIO2. QVS-LIVO achieved lower horizontal RMSE than all four external baselines on both sequences.

ORB-SLAM2 and ORB-SLAM3 achieved RMSEs of 1.197\,m and 1.323\,m on Dataset\#1, increasing to 4.425\,m and 4.011\,m on Dataset\#2. LOAM recorded 5.682\,m and 7.036\,m, respectively. FAST-LIO2 achieved 0.987\,m and 2.933\,m, corresponding to relative error reductions of 7.8\% and 43.3\% with QVS-LIVO. The substantial error reduction on Dataset\#2 (43.3\% relative to FAST-LIO2) highlights the benefit of selectively incorporating complementary visual constraints over long-route off-road traversal.

\subsubsection{Point-Cloud Consistency}
\label{sec:pointcloud-consistency}

Under normal input, the median local plane thickness P90 values for LIO, Front VIO, Joint VIO and QVS-LIVO were 47.554\,mm, 44.778\,mm, 48.727\,mm and 45.073\,mm, respectively. Achieving a P90 thickness close to Front VIO (45.073\,mm versus 44.778\,mm) and lower than LIO and Joint VIO indicates that QVS-LIVO maintains local surface consistency during view switching.

\begin{figure*}[!t]
\centering
\includegraphics[width=\textwidth]{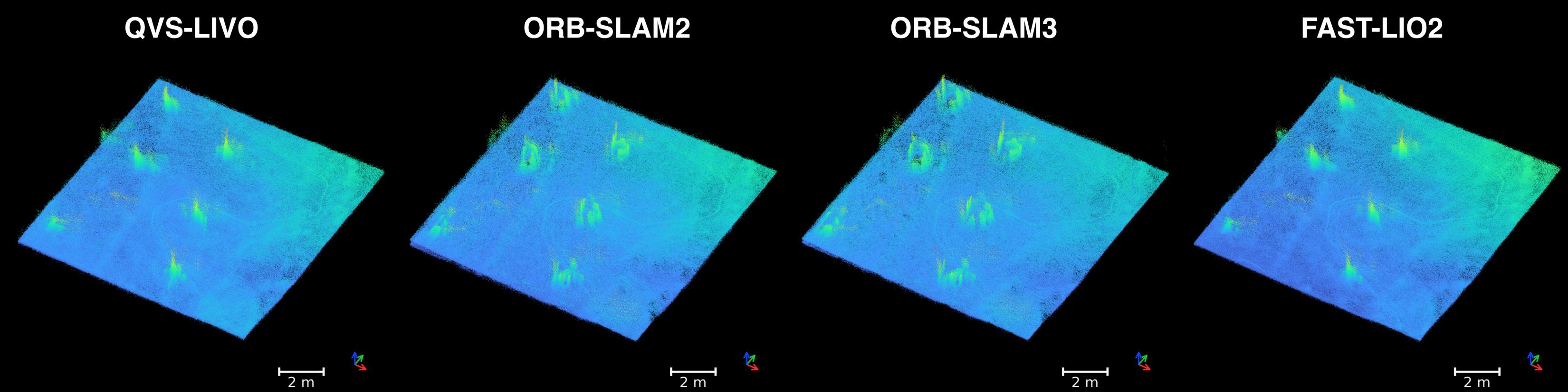}
\caption{Accumulated point clouds from QVS-LIVO, ORB-SLAM2, ORB-SLAM3 and FAST-LIO2 (left to right). Each panel includes a 2 m scale bar.}
\label{fig:pointcloud-accumulation}
\end{figure*}

Figure~\ref{fig:pointcloud-accumulation} shows accumulated point clouds from QVS-LIVO, ORB-SLAM2, ORB-SLAM3 and FAST-LIO2. Figure~\ref{fig:pointcloud-heatmap} projects the horizontal trajectory MSE and MAE onto the accumulated point-cloud footprint to illustrate how pose drift spatially distributes along the traversal route.

\afterpage{%
\begin{figure*}[!t]
\centering
\includegraphics[width=\textwidth]{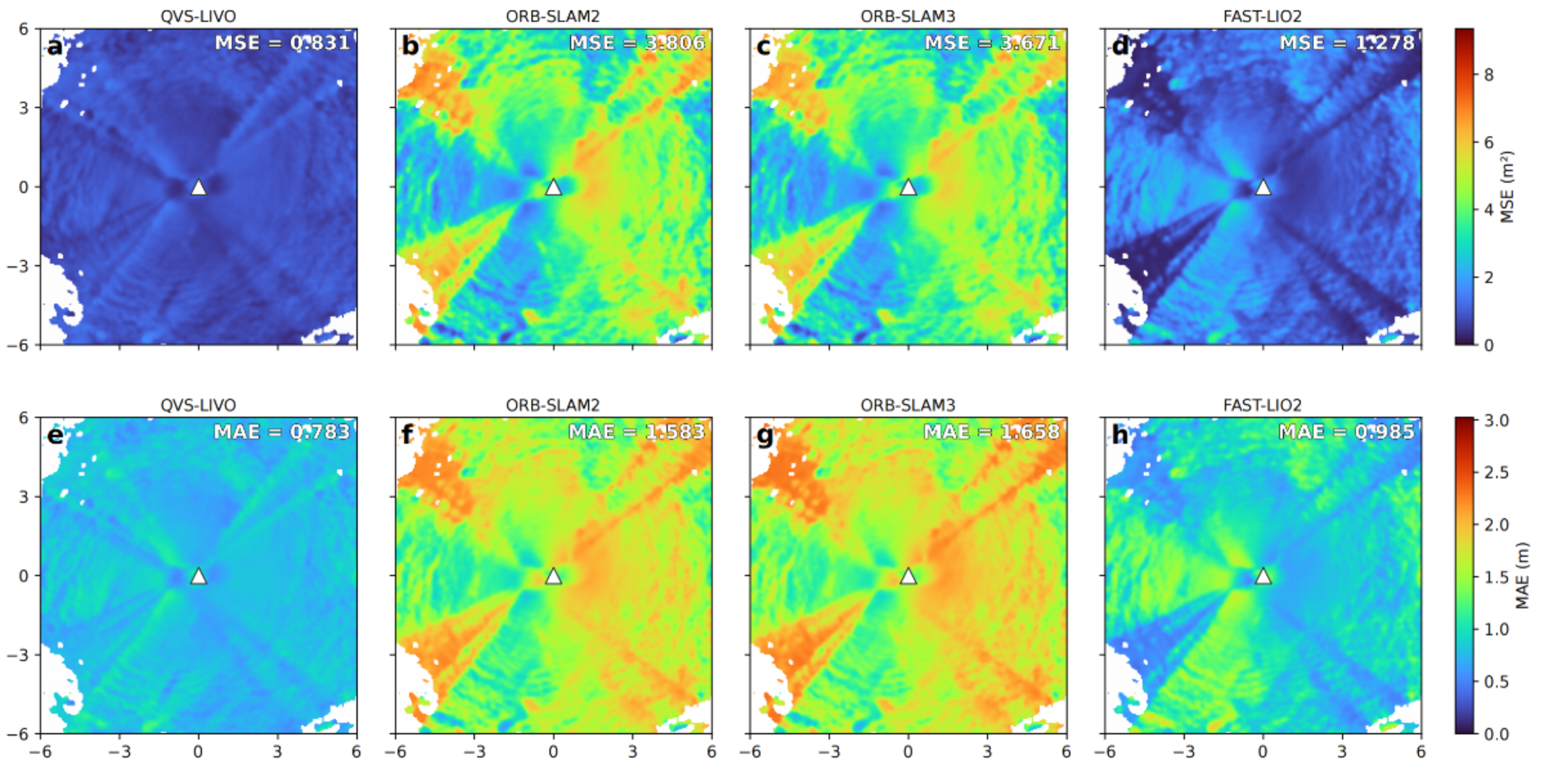}
\caption{Spatial visualization of horizontal trajectory errors for QVS-LIVO, ORB-SLAM2, ORB-SLAM3 and FAST-LIO2 (left to right): MSE (top, $\mathrm{m}^2$) and MAE (bottom, m). Errors relative to the reference positions are mapped using the point-cloud distribution. White regions indicate insufficient point-cloud coverage, triangles mark the vehicle center and forward direction. Coordinates are vehicle-centered (m), with forward upward and right to the right. Color scales are shared within each row.}
\label{fig:pointcloud-heatmap}
\end{figure*}
}

\subsection{Camera Fault Response}
\label{sec:fault-results}

Table~\ref{tab:faults} reports localization errors and trajectory availability for the 12 single-camera fault configurations on Dataset\#2, alongside normal-input baselines. Following Section~\ref{sec:controlled-faults}, the tests examined input interruption, invalid image content and temporal offsets during the source-time interval \([105,120)\,\mathrm{s}\). Lidar, IMU, depth and the remaining camera inputs were unchanged.

\subsubsection{RGB Dropout}
Camera dropout revealed differences in whether the evaluated configurations continued producing trajectories. Dropout of any of the four cameras caused processing backlog and trajectory termination in the evaluated Joint VIO implementation. For fixed-front-view FAST-LIVO2, front-camera dropout increased horizontal RMSE from 5.83\,m to 14.40\,m. ARTI-VIO increased from 7.27\,m to 17.42--29.24\,m across the four dropout configurations. QVS-LIVO maintained evaluable trajectories in all four configurations, with RMSEs of 1.76--2.21\,m. These results support the importance of handling individual channel interruptions when combining multiple visual streams, as discussed in Section~\ref{sec:internal-results}.

\subsubsection{Black Frames}
Black-frame input preserved image-message arrivals while removing image texture. Joint VIO achieved horizontal RMSEs of 2.08--2.39\,m. QVS-LIVO had lower errors for the front-, rear- and right-camera faults, at 1.75\,m, 1.86\,m and 1.94\,m, respectively. The two methods tied at 2.09\,m for the left-camera fault at the reported precision. ORB-SLAM2 and ORB-SLAM3 reported persistent tracking loss or invalid trajectory timestamps under black-frame input. Their left- and right-camera controls already exhibited these failure states under normal input. The QVS-LIVO results are consistent with observation-health screening before visual correction, as described in Section~\ref{sec:visual-selection}.

\subsubsection{Timestamp Shifts}
The \(+60\,\mathrm{ms}\) timestamp shift perturbed temporal correspondence with lidar and IMU measurements while preserving the original image content. For front-camera ORB-SLAM2, RMSE increased from 4.43\,m under normal input to 58.70\,m. Joint VIO achieved RMSEs of 2.14--3.25\,m, while QVS-LIVO achieved 1.54--2.23\,m across the four camera configurations. Although Joint VIO achieved a marginally lower RMSE under the left-camera timestamp shift (2.14\,m versus 2.23\,m), QVS-LIVO outperformed or matched Joint VIO across all remaining black-frame and timestamp-shift cases while avoiding the trajectory halts observed under camera dropout.

Across all 12 fault configurations, QVS-LIVO maintained complete trajectories with horizontal RMSEs of 1.54--2.23\,m (23.9\%--47.4\% lower than the camera-independent FAST-LIO2 baseline of 2.93\,m), demonstrating that selective visual updates coupled with LIO fallback effectively contain single-channel degradation.

\afterpage{%
\begin{sidewaystable*}[p]
\centering
\caption{Camera-fault results on Dataset\#2: horizontal RMSE (m) and the recorded source-time offset $T_{\mathrm{in}}$ (ms), defined below. Faults span $[105,120)$ s. ORB-SLAM2/3 use the affected camera.}
\label{tab:faults}
\fontsize{9.5}{11.5}\selectfont

\setlength{\tabcolsep}{4pt}
\renewcommand{\arraystretch}{1.65}
\begin{adjustbox}{max width=\linewidth}
\begin{tabular*}{\linewidth}{@{\extracolsep{\fill}}ll*{7}{cc}@{}}
\toprule
Camera & Condition & \multicolumn{2}{c}{QVS-LIVO} & \multicolumn{2}{c}{FAST-LIVO2~\citep{Zheng2025FastLIVO2}} & \multicolumn{2}{c}{Joint VIO} & \multicolumn{2}{c}{FAST-LIO2~\citep{Xu2022FastLIO2}} & \multicolumn{2}{c}{ORB-SLAM2~\citep{MurArtal2017ORBSLAM2}} & \multicolumn{2}{c}{ORB-SLAM3~\citep{Campos2021ORBSLAM3}} & \multicolumn{2}{c}{ARTI-VIO~\citep{Feng2026ARTIVIO}} \\
\cmidrule(lr){3-4}\cmidrule(lr){5-6}\cmidrule(lr){7-8}\cmidrule(lr){9-10}\cmidrule(lr){11-12}\cmidrule(lr){13-14}\cmidrule(lr){15-16}
 & & RMSE & $T_{\mathrm{in}}$ & RMSE & $T_{\mathrm{in}}$ & RMSE & $T_{\mathrm{in}}$ & RMSE & $T_{\mathrm{in}}$ & RMSE & $T_{\mathrm{in}}$ & RMSE & $T_{\mathrm{in}}$ & RMSE & $T_{\mathrm{in}}$ \\
\midrule
Baseline & Dataset\#2 & \textbf{1.66} & - & 5.83 & - & 3.31 & - & 2.93 & - & 4.43 & - & 4.01 & - & 7.27 & - \\
\midrule
\multirow{4}{*}{Front} & Normal input & 1.66 & - & - & - & 3.31 & - & 2.93 & - & 4.43 & - & 4.01 & - & 7.27 & - \\
 & Dropout & \textbf{1.76} & 103.14 & 14.40 & 1037.30 & TH & 103.14 & - & - & TL & 3.06 & IT & 3.06 & 29.24 & 36.41 \\
 & Black frames & \textbf{1.75} & 103.14 & 1.94 & 3.06 & 2.08 & 370.04 & - & - & TL & 3.06 & IT & 3.06 & 7.10 & 3.06 \\
 & Offset +60 ms & \textbf{2.09} & 69.78 & 2.17 & 36.41 & 2.58 & 136.50 & - & - & 58.70 & 36.41 & 4.00 & 36.41 & 6.47 & 36.41 \\
\midrule
\multirow{4}{*}{Left} & Normal input & 1.66 & - & - & - & 3.31 & - & 2.93 & - & TL & - & IT & - & 7.27 & - \\
 & Dropout & \textbf{2.06} & 100.29 & - & - & TH & 100.29 & - & - & TL & 0.21 & IT & 0.21 & 20.56 & 0.21 \\
 & Black frames & 2.09 & 66.93 & - & - & \textbf{2.09} & 33.57 & - & - & TL & 0.21 & IT & 0.21 & 5.37 & 0.21 \\
 & Offset +60 ms & 2.23 & 66.93 & - & - & \textbf{2.14} & 100.29 & - & - & TL & 33.57 & IT & 33.57 & 6.64 & 0.21 \\
\midrule
\multirow{4}{*}{Rear} & Normal input & 1.66 & - & - & - & 3.31 & - & 2.93 & - & 53.07 & - & 6.94 & - & 7.27 & - \\
 & Dropout & \textbf{2.10} & 132.96 & - & - & TH & 66.26 & - & - & TL & 32.88 & IT & 32.88 & 17.42 & 32.88 \\
 & Black frames & \textbf{1.86} & 66.26 & - & - & 2.39 & 66.26 & - & - & TL & 32.88 & IT & 32.88 & 7.72 & 32.88 \\
 & Offset +60 ms & \textbf{2.15} & 132.96 & - & - & 3.25 & 66.26 & - & - & 53.57 & 66.26 & 6.12 & 66.26 & 4.68 & 32.88 \\
\midrule
\multirow{4}{*}{Right} & Normal input & 1.66 & - & - & - & 3.31 & - & 2.93 & - & TL & - & IT & - & 7.27 & - \\
 & Dropout & \textbf{2.21} & 86.94 & - & - & TH & 153.66 & - & - & TL & 20.22 & IT & 20.22 & 26.93 & 53.57 \\
 & Black frames & \textbf{1.94} & 287.11 & - & - & 2.14 & 20.22 & - & - & TL & 20.22 & IT & 20.22 & 6.91 & 20.22 \\
 & Offset +60 ms & \textbf{1.54} & 86.94 & - & - & 2.37 & 20.22 & - & - & TL & 53.57 & IT & 53.57 & 10.96 & 20.22 \\
\bottomrule
\end{tabular*}
\end{adjustbox}
\par\vspace{12pt}\noindent
\begin{minipage}{\linewidth}

\fontsize{8.5}{10.5}\selectfont

$T_{\mathrm{in}}$: source-time offset of the first recorded post-fault tracking-recovery metric. TL: persistent tracking loss, TH: processing backlog and trajectory halt, IT: invalid trajectory timestamps, -: unavailable or not applicable.\\[3pt]
The top row uses the existing Dataset\#2 baselines. Front-camera ORB values repeat that baseline, other ORB normal-input values use the corresponding cameras. FAST-LIO2 is camera-independent. RMSE uses available trajectory timestamps and the C430i GNSS/IMU position reference. FAST-LIVO2 denotes Front VIO. ARTI-VIO is the four-camera RGB-D+IMU variant.
\end{minipage}
\end{sidewaystable*}
}

\afterpage{\afterpage{%
\begin{figure*}[!t]
\centering
\includegraphics[width=0.49\textwidth]{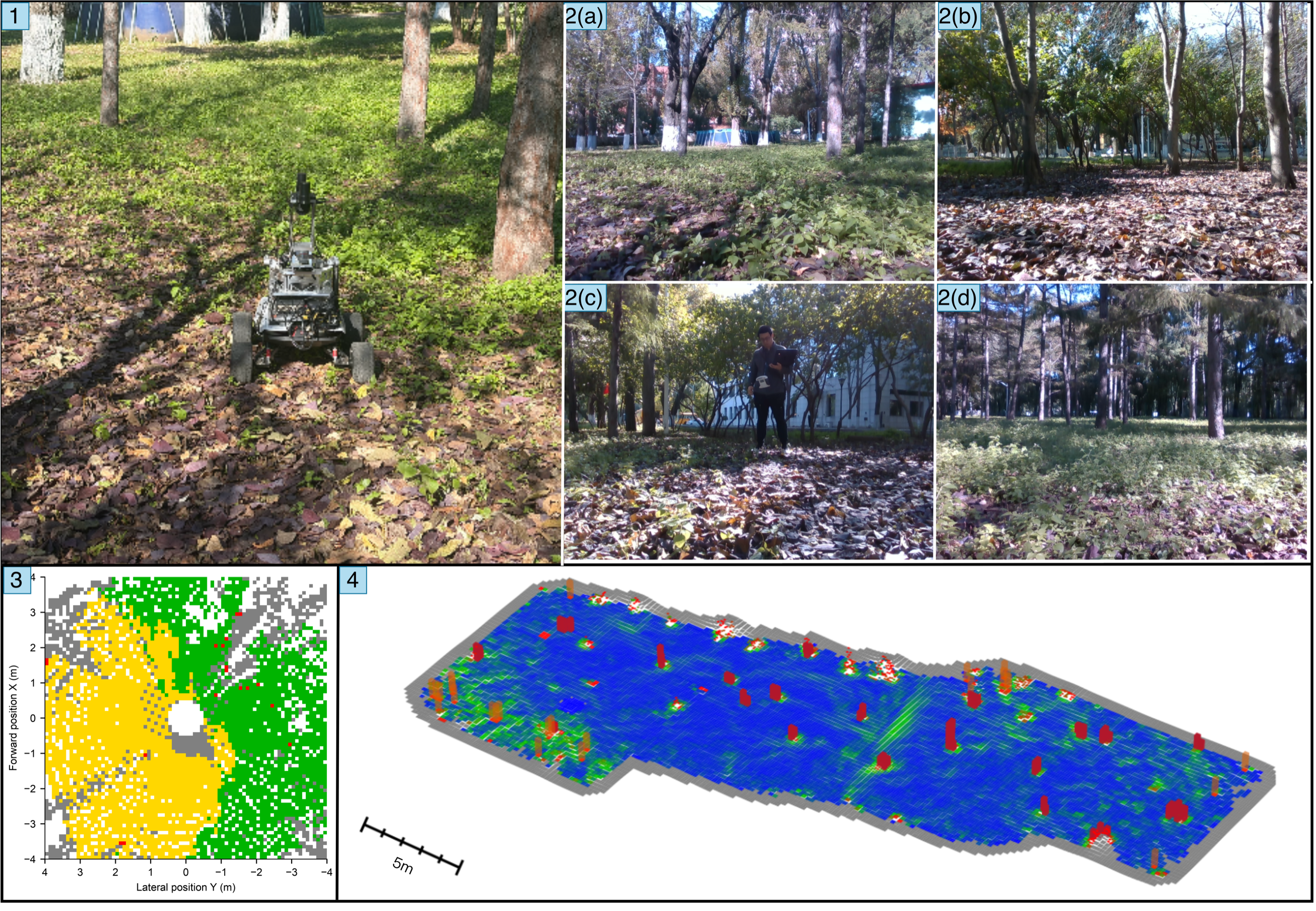}\hfill
\includegraphics[width=0.49\textwidth]{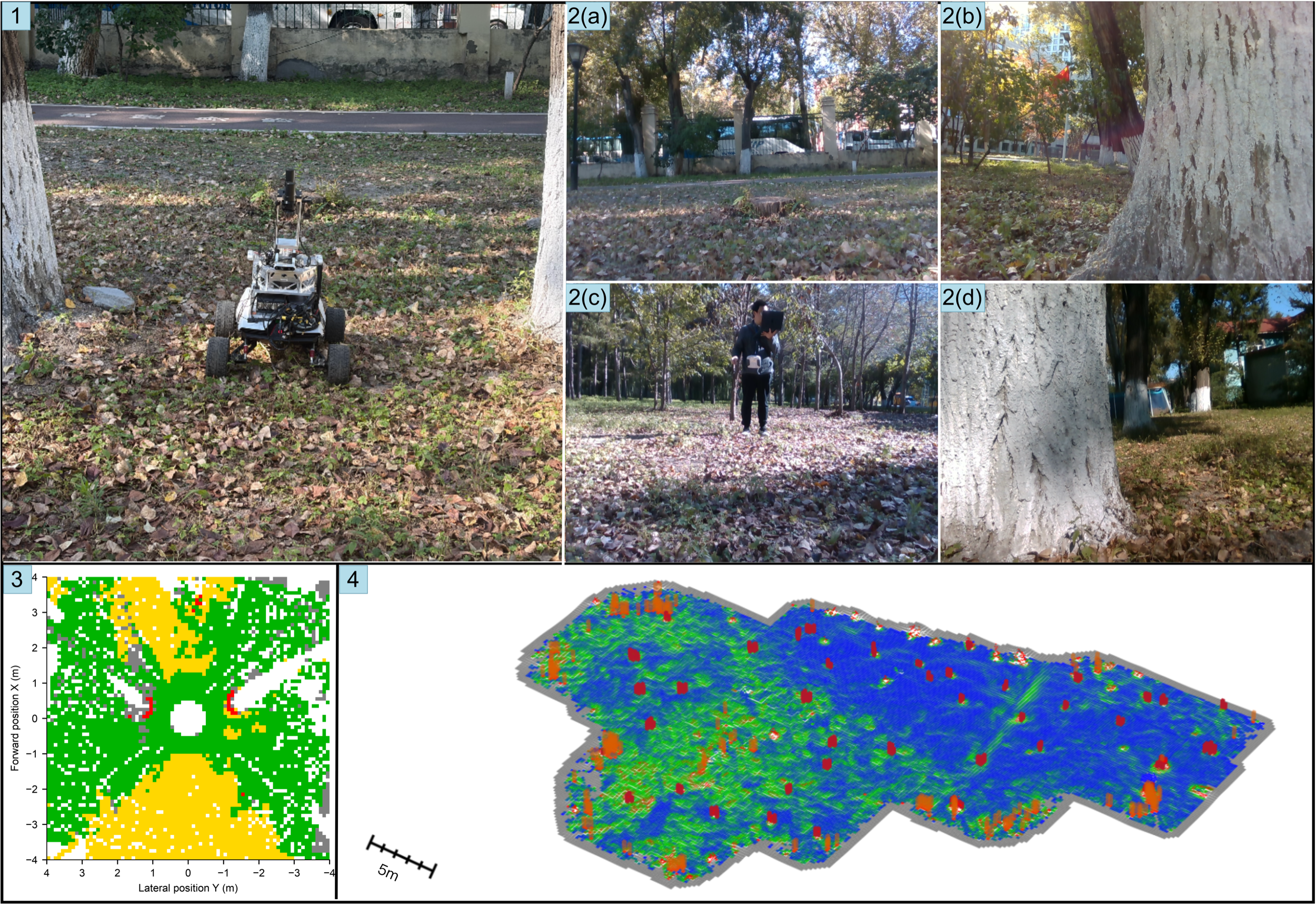}
\caption{Real-time semantic--geometric mapping during onboard deployment in two woodland scenes. In each column, (1) shows a third-person view of the robot in the field, and 2(a--d) show first-person views from the four cameras. Panels (3) and (4) show the Local Risk BEV and the Global Semantic--Terrain Risk Map with obstacle instance voxels.}
\label{fig:onboard-deployment}
\end{figure*}
}}

\subsection{Semantic--Geometric Risk Mapping Results}
\label{sec:mapping-results}

The final fine-tuned TraverNet checkpoint achieved a four-class mean intersection over union (mIoU) of 82.29\% on the 11 local validation images and 70.37\% on 962 GOOSE validation images~\citep{Mortimer2024GOOSE}. Local class IoUs were 89.77\%, 80.56\%, 94.14\% and 64.68\% for grass, road, tree and other, respectively. 

Mapping on the collected field sequences produced three complementary representations: a Local Risk BEV, a Global Semantic--Terrain Risk Map (2.5D) and Obstacle Instance Voxels. The local map represents surface classes, continuous risk and confirmation states around the robot. The global map accumulates terrain geometry, semantic attributes and risk estimates along the route, while the instance voxels retain the three-dimensional structure of above-ground obstacles.

These outputs incorporate semantic observations from all available camera views, independently of localization-view selection. The mapping pipeline projects both newly registered points and accumulated ground points into images, allowing subsequent observations to add semantic information to previously observed geometry.

In the Dataset\#1 scene statistics, mean risk values for \texttt{road}, \texttt{grass}, \texttt{other} and \texttt{tree} were 0.159, 0.287, 0.660 and 0.741, respectively. Road and grass had lower mean risks than other and tree. This progression aligns with physical traversability expectations, reflecting the consistent coupling between measured 2.5D terrain geometry and bounded appearance priors.

The map outputs retain continuous risk and confirmation status separately, distinguishing estimated hazard magnitude from the evidence supporting its confirmation. Regions without valid geometric observations remain unknown, preserving the distinction between unobserved terrain and assessed risk.

\subsection{Runtime Performance}
\label{sec:runtime-results}

Figure~\ref{fig:onboard-deployment} illustrates onboard deployment of QVS-LIVO RiskMap, showing the field scenes, camera observations and corresponding local and global maps.

\subsubsection{Semantic Inference and Projection}
\label{sec:semantic-runtime}

During workstation replay, standalone semantic inference times were 4.092\,ms at P50 and 4.671\,ms at P95. The aggregate semantic output rate across the four camera streams approached 10\,Hz. Across the two datasets, valid semantic projection output rates ranged from 8.16 to 9.04\,Hz, with timestamp-pairing rates of 88.73\%--98.75\%.

\subsubsection{Map Timing and Output Rates}
\label{sec:deployment-latency}

Table~\ref{tab:deployment-latency} summarizes map timing measurements and publication rates in onboard deployment and workstation replay. Onboard measurements used an AGX Orin, with observation windows of 30\,s and 180\,s for the local and global maps, respectively. Workstation timing measurements used a separate 145.24\,s Dataset\#1 sequence on an Intel Core i7-14700KF and NVIDIA RTX 4090. The global replay pipeline used the same resource allocation configuration as the robot and published at 2\,Hz. Its timing covered snapshot construction, derived-layer computation and publication calls. Output rates were computed from actual map outputs.

\par\medskip\noindent
\begin{minipage}{\columnwidth}
\centering
\captionof{table}{Map timing measurements and output rates in onboard deployment and workstation replay.}
\label{tab:deployment-latency}
\scriptsize
\setlength{\tabcolsep}{3pt}
\renewcommand{\arraystretch}{1.16}
\begin{tabular}{@{}llrrr@{}}
\toprule
Experiment & Map & Rate (Hz) & Median (ms) & P95 (ms) \\
\midrule
On-robot & Local BEV & 9.94 & 32.4 & 37.9 \\
On-robot & Global map & 2.00 & 486.4 & 516.1 \\
Replay & Local BEV & 10.00 & 13.2 & 22.5 \\
Replay & Global map & 2.00 & 75.9 & 124.0 \\
\bottomrule
\end{tabular}
\par\vspace{3pt}

\end{minipage}
\par\medskip

On the AGX Orin, the Local Risk BEV updated at 9.94\,Hz with a median processing time of 32.4\,ms. The global map published at 2.00\,Hz, with a recorded median time of 486.4\,ms. Workstation replay yielded 1401 local and 279 global map samples. Their output rates were 10.00\,Hz and 2.00\,Hz, with median processing times of 13.2\,ms and 75.9\,ms, respectively. These measurements demonstrate onboard operation at approximately 10\,Hz for local risk-map updates and 2\,Hz for global map publication.

\subsection{Discussion}
\label{sec:discussion}

\subsubsection{Decoupled View Utilization for Localization and Mapping}
\label{sec:discussion-view-utilization}

The experimental results highlight a fundamental divergence between how multi-camera observations benefit state estimation versus environmental mapping. In our internal evaluations, QVS-LIVO achieved the lowest horizontal RMSE while retaining over three times as many colored map points as the single-camera baseline (Front VIO) on Dataset\#2. Notably, although Joint VIO yielded the highest raw visual-tracking availability ($\eta_{\mathrm V}$) under the at-least-one-camera criterion, incorporating all available views simultaneously resulted in degraded localization accuracy compared to both QVS-LIVO and the camera-independent LIO baseline. This discrepancy shows that tracking availability does not automatically translate into superior pose constraints; degraded or geometrically redundant views may provide limited benefit to the lidar--inertial posterior. Separating selective visual state updates from all-view semantic aggregation addresses these different requirements, supporting pose estimation alongside broader spatial map coverage.

The controlled fault-injection experiments further underscore the necessity of active channel management in multi-camera fusion. Merely aggregating multiple streams does not guarantee resilience against single-channel anomalies. When subjected to single-camera dropouts, the Joint VIO baseline suffered from synchronization backlogs and premature trajectory termination. Conversely, QVS-LIVO maintained evaluable trajectories across all 12 fault configurations (dropout, black frames, and $+60\,\mathrm{ms}$ timestamp offsets) by combining read-only pretracking with stateful channel isolation and post-solve acceptance checks. Whenever visual candidates became unreliable or temporally inconsistent, the estimator retained the shared LIO posterior rather than committing the visual correction.

From a computational and system-design perspective, this decoupled architecture introduces specific operational trade-offs. Restricting the iterative photometric update in the ESIKF to at most one winning camera bounds the optimization cost, supporting onboard processing on the AGX Orin. However, lightweight multi-view pretracking and four-camera semantic inference still impose baseline computational overhead, while asynchronous multi-view projection demands strict timestamp pairing to avoid spatial ghosting. Furthermore, the dual-scale schedule updates the robot-centric Local Risk BEV at 9.94\,Hz (32.4\,ms median processing time on the AGX Orin) while publishing the Global Semantic--Terrain Risk Map at 2.00\,Hz. This allocation supports frequent local risk updates alongside longer-horizon environmental modeling.

\subsubsection{Asymmetric Fusion and Evidence-Based Risk Confirmation}
\label{sec:discussion-risk-confirmation}

In unstructured off-road terrain, maintaining separate representations for continuous risk magnitude, binary confirmation status, and unknown space provides richer semantics for downstream navigation than a single collapsed costmap. Continuous risk captures graduated hazard severity by blending vehicle-scale geometry with appearance priors, whereas confirmation status reflects whether the accumulated spatiotemporal evidence is sufficient to warrant a hard traversal constraint. Preserving this distinction allows uncertain, high-risk candidates to be flagged for cautious inspection without prematurely blocking the planner, while explicitly retaining unknown states prevents unobserved blind zones from being misinterpreted as safe terrain.

Furthermore, our hierarchical fusion and confirmation rules enforce an intentional asymmetry between geometric and semantic authority. Because visual segmentation in field environments is prone to transient misclassifications under canopy shadows or domain shifts, appearance priors are restricted to bounded risk adjustments and candidate vetoes on moderate slopes or steps. Crucially, semantic evidence can neither override direct geometric occupancy within the vehicle's passage band nor independently synthesize hard obstacles. By confirming high-confidence geometric obstacles immediately while requiring spatiotemporal persistence for slope- and step-induced hazards, the pipeline is designed to suppress false alarms caused by sparse lidar edge outliers or transient vegetation observations.

Nevertheless, spatiotemporal confirmation inherently trades reaction latency and minimum detectable hazard size for false-positive suppression. Requiring connected spatial support and multi-cycle temporal voting can briefly delay the confirmation of emerging terrain discontinuities or filter out thin, spatially isolated hazards whose footprint falls below the grid support threshold. The optimal operating point along this trade-off curve depends on grid resolution, update frequency, and traversal speed.

\section{Conclusion}
\label{sec:conclusion}

This paper presented QVS-LIVO RiskMap, a multi-camera localization and dual-scale risk-mapping framework tailored for unstructured off-road environments. The core architectural insight is to decouple selective visual state estimation from all-view semantic mapping while anchoring both branches to shared pose estimates and registered point clouds. In the localization front end, read-only multi-view pretracking on a shared LIO posterior evaluates both observation health and directional geometric complementarity. Governed by stateful channel memory, switching hysteresis, and post-solve acceptance checks, the estimator admits at most one reliable view for direct photometric correction and safely retains the LIO prior whenever visual updates are degraded or inconsistent. Concurrently, the mapping back end aggregates semantic streams across all available cameras into a world-anchored 2.5D grid, enforcing vehicle-scale geometric authority and spatiotemporal confirmation to generate a Local Risk BEV, a Global Semantic--Terrain Risk Map, and 3D Obstacle Instance Voxels.

Extensive field evaluations and controlled fault-injection benchmarks validate the proposed framework across localization, mapping, and onboard deployment. Across two challenging off-road sequences, QVS-LIVO achieved horizontal trajectory RMSEs of 0.910\,m and 1.664\,m---reducing long-route drift by up to 43.3\% relative to FAST-LIO2---while retaining 2.24 to 3.02 times as many colored map points as fixed single-view fusion. Tracking diagnostics and 12 single-camera fault configurations (spanning stream dropout, black frames, and $+60\,\mathrm{ms}$ timestamp shifts) confirmed that selective single-view updates outperform indiscriminate multi-camera fusion, sustaining uninterrupted trajectories with RMSEs of 1.54--2.23\,m where baseline multi-view estimators suffered from severe drift or premature termination. Deployed onboard an NVIDIA Jetson AGX Orin, the complete pipeline operated in real time, updating local and global risk maps at 9.94\,Hz (32.4\,ms median latency) and 2.00\,Hz, respectively.

While QVS-LIVO RiskMap demonstrates robust localization and real-time dual-scale mapping in challenging field conditions, several avenues remain for future investigation. First, conducting component-level ablations under strictly equalized onboard compute budgets will further quantify the individual gains of health evaluation, directional complementarity, and hysteretic switching. Second, extending the temporal consistency checks to actively estimate and compensate for online inter-camera clock drift could improve resilience under persistent synchronization faults, such as the timestamp offsets examined in this study. Finally, future work will focus on benchmarking the risk representations against dense ground-truth traversability annotations and integrating the dual-scale maps into a closed-loop motion planner for autonomous off-road navigation.

\bmsubsection*{Author Contributions (CRediT)}
\textbf{Zihan Pan}: Conceptualization, Methodology, Investigation
, Data curation, Writing - original draft.
\textbf{Jinghan Wang}: Validation, Visualization, Writing – review \& editing.
\textbf{Tianchen Liu}:Supervision, Writing – review \& editing.
\textbf{Hang Li}:Project administration.
\textbf{Feng Cheng}:Project administration.
\textbf{Gaoliang Peng}:Supervision, Funding acquisition.

\bmsubsection*{Conflicts of Interest}

The authors declare no conflicts of interest.

\end{document}